\documentclass[10pt,twocolumn,letterpaper]{article}
\usepackage[pagenumbers]{cvpr}

\usepackage[accsupp]{axessibility}
\usepackage{multirow}
\usepackage{ifthen}
\usepackage{verbatim}
\usepackage{subcaption}
\usepackage[dvipsnames]{xcolor}
\usepackage{tabu}
\usepackage{amssymb}

\usepackage{eso-pic}
\usepackage{pdfpages}
\usepackage{multido}

\usepackage{float}
\usepackage{pifont}
\usepackage{color, colortbl}

\DeclareCaptionFormat{teaser}{%
    \textbf{#1#2}#3
}

\newcommand{\cmark}{\ding{51}}%
\newcommand{\xmark}{\ding{55}}%

\definecolor{cvprblue}{rgb}{0.21,0.49,0.74}
\usepackage[pagebackref,breaklinks,colorlinks,citecolor=cvprblue]{hyperref}

\newcommand{\notsosmall}{\fontsize{10.5pt}{12pt}\selectfont}

\def\paperID{93}
\def\confName{3DV\xspace}

\definecolor{somegray}{rgb}{0.5, 0.5, 0.5}
\newcommand{\darkgrayed}[1]{\textcolor{somegray}{#1}}
\makeatletter
\newcommand*\titleheader[1]{\gdef\@titleheader{#1}}
\AtBeginDocument{%
  \let\st@red@title\@title
  \def\@title{%
    \vskip-3em
    \bgroup\normalfont\large\centering\@titleheader\par\egroup
    \vskip1.5em\st@red@title}
}
\makeatother

\titleheader{\darkgrayed{This paper has been accepted for publication at the \\
IEEE International Conference on 3D Vision (3DV), Davos, 2024.
\copyright IEEE}}

\title{Revisiting Depth Completion from a Stereo Matching Perspective \\ for Cross-domain Generalization}

\author{Luca Bartolomei$^{*,\dagger}$ \hspace{0.7cm} Matteo Poggi$^{*,\dagger}$ \hspace{0.7cm} Andrea Conti$^\dagger$ \hspace{0.7cm} Fabio Tosi$^\dagger$ \hspace{0.7cm} Stefano Mattoccia$^{*,\dagger}$ \\
\notsosmall $^*$Advanced Research Center on Electronic System (ARCES) \\ 
\notsosmall $^\dagger$Department of Computer Science and Engineering (DISI) \\
\notsosmall University of Bologna, Italy \\
{\tt\small\{luca.bartolomei5, m.poggi, andrea.conti35, fabio.tosi5 stefano.mattoccia\}@unibo.it} \\
\small\url{https://vppdc.github.io/}
}

\begin{document}

\twocolumn[{%
\maketitle
\vspace{-1cm}
   \begin{center}    
   \includegraphics[width=1\textwidth]{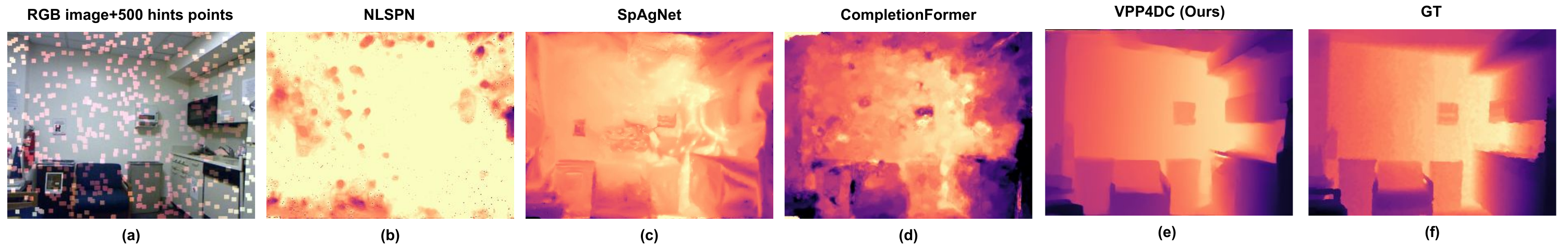}  
   \vspace{-0.7cm}
    \captionof{figure}{\textbf{Synth-to-real generalization.} 
    Given an NYU Depth V2 \cite{nyudepthv2} frame and 500 sparse depth points (a), our framework with RAFT-Stereo \cite{lipson2021raft} trained only on the Sceneflow \cite{MIFDB16} synthetic dataset (e) outperforms the generalization capability of state-of-the-art depth completion networks NLSPN \cite{park2020non} (b), SpAgNet \cite{Conti_2023_WACV} (c), and CompletionFormer \cite{zhang2023completionformer} (d) -- all trained on the same synthetic dataset.}
  \label{fig:teaser}
  \end{center}
  \vspace{0.4cm}
}]

\begin{abstract}
This paper proposes a new framework for depth completion robust against domain-shifting issues. It exploits the generalization capability of modern stereo networks to face depth completion, by processing fictitious stereo pairs obtained through a virtual pattern projection paradigm.
Any stereo network or traditional stereo matcher can be seamlessly plugged into our framework, allowing for the deployment of a virtual stereo setup that is future-proof against advancement in the stereo field. Exhaustive experiments on cross-domain generalization support our claims. Hence, we argue that our framework can help depth completion to reach new deployment scenarios.
\end{abstract}    
\section{Introduction}

Perceiving depth is of utmost importance for several computer vision tasks, such as autonomous driving, robotics, and augmented reality, to name a few of them. Different camera setups exist for this purpose in the literature, ranging from single to multiple imaging systems, and deep network approaches dominate these fields. An alternative strategy to infer depth in practical applications relies on deploying active depth sensors based on LiDAR or Time-of-Flight technologies, despite they feature much lower resolution than conventional cameras. 
To overcome the limited depth density of active depth sensors, they are frequently coupled with higher-resolution imaging devices in stereo or monocular camera setups to achieve dense depth maps at higher resolution. The monocular setup, consisting of a conventional camera registered with an active depth sensor, has received much attention recently due to the minimalist setup needed and the outstanding results achieved by learning-based methods to tackle this task, known in the literature as depth completion. 
Despite these achievements, state-of-the-art networks struggle with out-of-domain data distribution, making their practical deployment challenging. In fact, these networks are tightly constrained to the depth sensor/data distribution and environments seen during training \cite{hu2022deep}. Nevertheless, increasing generalization robustness has been often overlooked and consequently scarcely investigated in the literature, with most studies limiting the training and testing of depth completion solutions over a single, very specific domain -- either indoor \cite{nyudepthv2} or outdoor \cite{kittidc}.

Purposely, this paper aims to fill this gap by studying this issue deeply and tackling it from a different perspective. Inspired by the outstanding generalization capability of modern stereo networks \cite{teed2020raft,Xu_2023_CVPR}, we cast depth completion as if it were a stereo problem through a depth-based virtual pattern projection paradigm. 
In contrast to most depth completion approaches conceived to densify the input sparse depth seeds according to the image content and monocular cues only, our strategy treats depth completion as a correspondence problem through existing stereo matchers. This is achieved by processing virtual stereo pairs characterized by less domain-specific features, enabling much higher robustness to out-of-domain issues.

Our main contributions can be summarized as follows:

\begin{itemize}
    \item We cast depth completion as a virtual stereo correspondence problem, where two appropriately patterned virtual images enable us to face depth completion with robust stereo-matching algorithms or networks.
    \item Extensive experimental results with multiple datasets and networks demonstrate that our proposal vastly outperforms state-of-the-art concerning generalization capability, as shown in Fig. \ref{fig:teaser}, and performs comparably in-domain.  
\end{itemize}
\section{Related Work}
This section reviews the depth completion and stereo matching literature since both relate to our proposal.

\textbf{Depth Completion.} Given a sparse and potentially noisy depth map, gathered even from passive source \cite{void}, depth completion aims to predict missing values. The integration of an RGB camera provides a promising solution to address the challenges posed by pure sparse depth measurements, leading to more robust and reliable results.
A variety of traditional methods tackle this task via interpolation \cite{camplani2012efficient}, stochastic models \cite{shen2013layer}, morphological operators \cite{scpu} or geometric models \cite{zhao2021surface}.
Earlier deep learning approaches used convolutional neural networks (CNNs) \cite{ma2018sparse}, despite their suboptimal performance with sparse inputs. Some efforts, like \cite{kittidc}, attempted to handle  sparse measurements using sparsity-invariant CNNs. However, more recent advancements in guided spatial propagation-based networks have shown better results. For example, \cite{liu2017learning} proposed a network able to learn local affinities, which was further improved by \cite{cheng2018depth} and then by \cite{park2020non}. \cite{zhang2023completionformer} enhanced the latter approach using a jointly convolution-attention mechanism integrated into the encoder-decoder architecture.
Finally, regardless of the works above, \cite{choi2021stereodc} exploits 3D geometric cues recasting depth completion as a stereo matching problem. Their complex framework consists of: i) an image inpainting network to synthesize the missing stereo image by filling the warped sparse RGB image using the reference RGB image as context; ii) a depth-guided stereo matching network originally proposed by \cite{wang20193d}; iii) a depth refinement module.
While our research and \cite{choi2021stereodc} share some similarities, it is important to highlight the key differences. 
In their work, the inpainting module generates an image that may be ambiguous in uniform regions and repetitive patterns. In contrast, our approach focuses on extracting valuable information from sparse depth points using a virtual stereo pair with highly distinctive patterns. 
Additionally, the synthesis process itself is learned and thus suffers from a lack of generalization across domains or when dealing with very sparse data.
Instead, we leverage our virtual matching-friendly pattern to transform depth points into the image modality efficiently. 

\textbf{Stereo Matching.} Stereo matching aims at reconstructing a 2.5D scene from a rectified pair of images. Traditional methods \cite{SZELISKI_BOOK} rely on handcrafted algorithms, considering local \cite{Secaucus_1994_ECCV} and global information \cite{veksler2005stereo, yang2010constant, taniai2014graph, kolmogorov2004energy, boykov2001fast}. \cite{hirschmuller2007stereo} introduced a polynomial approximation method for efficient yet accurate stereo matching.
Recently, deep-learning solutions \cite{poggi2021synergies} outperformed conventional stereo on standard benchmarks \cite{zbontar2016stereo}. Initially, neural networks replaced some steps of the traditional pipeline. However, a paradigm shift has occurred with the emergence of end-to-end approaches. These approaches diverge based on their utilization of either a 2D encoder-decoder architecture \cite{MIFDB16, Liang_2018_CVPR, saikia2019autodispnet, yin2019hierarchical, Tankovich_2021_CVPR} or a 3D architecture that relies on feature cost volumes \cite{Kendall_2017_ICCV, chang2018psmnet, zhang2019ga, cheng2019learning, Shen_2021_CVPR}.
More recent works \cite{lipson2021raft, Xu_2023_CVPR}, instead, propose novel deep stereo networks that leverage the iterative refinement paradigm from the state-of-the-art optical flow network RAFT \cite{teed2020raft}, or rely on Vision Transformers \cite{xu2022unifying} to capture long-range contextual information.
Nevertheless, deep stereo methods often encounter challenges when estimating depth in unseen scenarios, and different methods have been proposed to address this issue \cite{zhang2019domaininvariant,cai2020matchingspace,aleotti2021neural,liu2022graftnet,chuah2022itsa,watson2020stereo,Tosi_2023_CVPR}. 
In tackling these challenges, other solutions have explored the integration of depth points obtained from external sensors (\eg LiDAR) to enhance depth estimation either by concatenating them as input of CNN-based architectures \cite{LIDARSTEREONET,choe2021volumetric, park2018high, zhang2020listereo, wang20193d} or by using them to guide the cost aggregation of existing cost volumes \cite{poggi2019guided, huang2021s3, zhang2022lidar, wang20193d}. In contrast, \cite{bartolomei2023ICCV} augments the given stereo pair to enhance RGB images, providing more discriminative information to the network and making it easier to solve the correspondence problem. Guided by the insights of this latter work in the stereo domain, we improve and extend the paradigm to the depth completion task. 

\begin{figure*}
    \centering
     \includegraphics[width=0.85\linewidth]{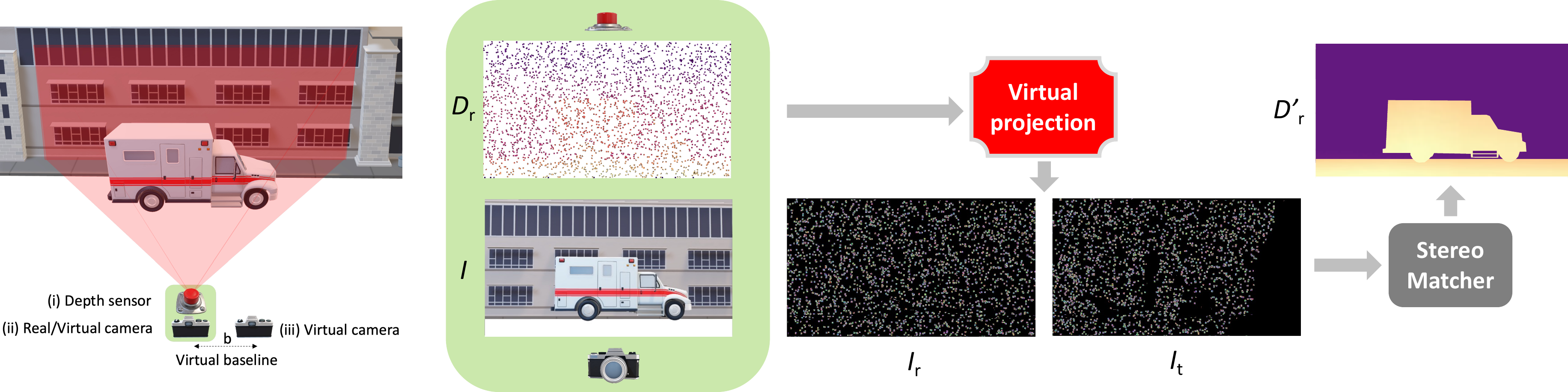}
     \vspace{-0.3cm}
     \caption{\textbf{Overview of the basic VPP4DC paradigm.} On the left, the proposed stereo setup we designed on top of the standard depth completion sensor suite enclosed in the green area. On the right, an outline of the proposed random projection that allows feeding a stereo matcher with a fictitious virtual patterned stereo pair and, optionally, an RGB image to tackle depth completion.}\vspace{-0.3cm}
     \label{fig:overview}
 \end{figure*}

\section{Virtual Pattern Projection for Depth Completion (VPP4DC)}

This section describes our approach for robustly facing depth completion, by deploying stereo matchers that process purely hallucinated image pairs generated according to the virtual pattern projection paradigm \cite{bartolomei2023ICCV} -- as if the scene were framed by two fictitious cameras and a pattern projector hitting the scene in sparse regions. The intuition behind this choice is to leverage the robustness of state-of-the-art stereo matchers \cite{lipson2021raft,Xu_2023_CVPR} at locating features across frames and domains to overcome the intrinsic limitations of conventional completion frameworks.

\subsection{Virtual Stereo Setup}

Given the standard setup for depth completion enclosed in the green area in Fig. \ref{fig:overview} -- consisting of a depth sensor (i) and an optional RGB camera (ii) -- our proposal casts the task as a stereo correspondence problem using a virtual stereo setup with two fictitious cameras, one in the same position as the actual RGB device if present (ii), and the other (iii) at a distance $b$, \ie the virtual stereo baseline. While the focal length $f$ of the virtual cameras is constrained by the depth sensor (i) or the RGB camera (ii), the virtual stereo baseline $b$ is a hyper-parameter.

We assume that the real RGB camera and the depth sensor are calibrated and we set the origin of the reference system in the camera. Therefore, we can project \cite{SZELISKI_BOOK} sparse depth points $\mathbf{Z}$ in the reference RGB camera view using the camera matrix $\mathbf{K}_r$ and the roto-translation $[\mathbf{R}_r|\mathbf{T}_r]$ between the depth sensor and the RGB camera:

\begin{equation}
    Z_r = \mathbf{K}_r \left[ \mathbf{R}_r | \mathbf{T}_r \right]  \mathbf{Z}
    \label{eq:reference_projection}
\end{equation}
where $Z_r$ is the sparse depth map projected into the reference image plane. The proximity of the depth sensor and RGB camera can reduce occlusion issues \cite{aconti2022lidarconf} when projecting, although they cannot be entirely avoided -- yet, can be easily identified and filtered out \cite{aconti2022lidarconf}.

Then, we place an additional target virtual camera sharing the same intrinsics $\mathbf{K}_r$ of the other virtual device at a horizontal distance to create a virtual baseline $b$.
Although we will stick to this setup, it is worth noting that the target virtual camera is not constrained to the horizontal axis. 

\subsection{Virtual Pattern Projection} 
\label{sec:vpp} 
In the outlined setup, we aim to project onto the two fictitious cameras appropriate virtual patterns coherent with the 3D structure of the scene framed by the depth sensor, as if a projector were present in the setup \cite{bartolomei2023ICCV}.
At first, the sparse depth points are converted to the disparity domain using the parameters of the virtual stereo rig \cite{SZELISKI_BOOK} as follows:

\begin{equation}
    D_r = \frac{b \cdot f}{Z_r}
    \label{eq:depth2disp}
\end{equation}
where $Z_r$ is the sparse depth map aligned with the reference image, $b$ is the virtual baseline, and $f$ is the focal length of the virtual cameras (the same as the RGB camera). $D_r$ is the sparse disparity map aligned with the reference virtual image $I_r$ and the RGB image $I$. 

Given our setup and the sparse depth points converted into disparity values, we can project the same pattern onto the fictitious reference $I_r$ and target $I_t$ cameras for each point $(x,y)$ with an available disparity value $d(x,y)$ in the reference image. It can be done by recalling \cite{SZELISKI_BOOK} that with a calibrated stereo system, the disparity $d(x,y)$ links one point $I_r(x,y)$ in the reference image with the corresponding $I_t(x',y)$ point in the target, with $x'=x-d(x,y)$. 
Once the two fictitious images have been generated, a stereo matcher processes them and produces a disparity map, that is then triangulated back into a densified depth map.

For projection: we manage real-valued disparities and occlusions, respectively by i) applying \emph{weighted splatting} in the target image and ii) reprojecting the foreground pattern on occluded regions, as in \cite{bartolomei2023ICCV}.
Independently of the pattern choice, discussed next, the process outlined is feasible only for a subset of the image points, and we set other points to a constant color (e.g., black in all our experiments). Therefore, from a different point of view, each fictitious camera gathers sparse content coherent with the 3D structure of the scene only where a fictitious virtual pattern projector sends its rays. 
Regarding the virtually projected patterns, we outline the two following strategies.

\textbf{RGB Projection.} We project onto the two fictitious images the same content $I(x,y)$ from the real camera, for each pixel with an available disparity value:  

\vspace{-0.3cm}\begin{equation}
    \begin{split}
        I_r(x,y) \leftarrow I(x,y), &\hspace{1cm}
        I_t(x',y) \leftarrow I(x,y), \\
        x'={} &x-d(x,y)
    \end{split} 
    \label{eq:rgb_pattern}
\end{equation}

\textbf{Random Pattern Projection.} Instead of warping the image content, we project more {\em matching-friendly} patterns. Following \cite{bartolomei2023ICCV}, we project coherently distinctive patterns onto the two fictitious cameras:

\vspace{-0.3cm}\begin{equation}
    \begin{split}
        I_r(x,y) \leftarrow \mathcal{P}, &\hspace{1cm}
        I_t(x',y) \leftarrow \mathcal{P}, \\
        x'={} &x-d(x,y)
    \end{split} 
    \label{eq:vpp_pattern}
\end{equation}
where operator $\mathcal{P}$ generates a random point-wise pattern applied coherently to both images. 
Fig. \ref{fig:overview} shows the output of this virtual projection strategy as an intermediate output of the VPP4DC module. 

On the one hand, compared to RGB Projection the random patterns are inherently less ambiguous by construction, for instance, in regions featuring a uniform texture. 
On the other hand, the sparse patterning prevents a complete awareness of the whole scene content for both strategies. However, such cue can be partially recovered from the RGB image if the stereo matcher can exploit image context \cite{lipson2021raft,Xu_2023_CVPR}.

\subsection{Additional Virtual Projection Strategies}

We extend the strategy outlined so far to i) increase pattern density according to the RGB content and ii) handle issues regarding the horizontal field of view in the stereo system. 

\begin{figure}[t]
    \centering
     \includegraphics[width=0.9\linewidth]{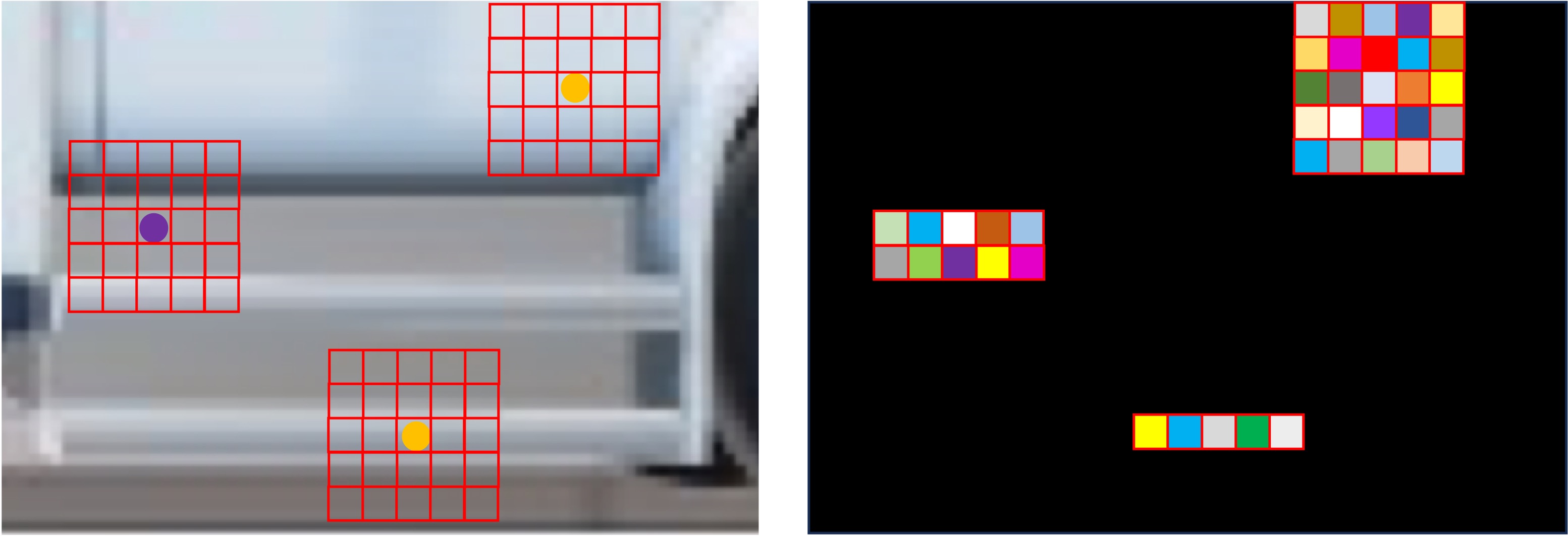}
     \vspace{-0.3cm}
     \caption{\textbf{Adaptive random patches.} The patch adapts to the image content; we pick a random color for each projected point. The figure shows only the fictitious patterned image $I_r$.}
     \label{fig:adaptive}
 \end{figure}

\subsubsection{Adaptive Patch-based Pattern Projection}

The basic point-wise patterning strategy can be extended to increase the pattern density, as proposed in \cite{bartolomei2023ICCV}, at nearby points simply by assuming the same disparity value locally. However, it can lead to the degradation of fine details as they can be lost adopting this method. Purposely, we exploit the RGB image $I$ to overcome this issue since it contains dense and meaningful cues about the scene. Specifically, we propose a heuristic inspired by the bilateral filter \cite{Bilateral_Filter} to adapt the shape of the patch and handle overlapping patches.

Given a fixed size patch $\mathcal{N}(x,y)$ centered on an available disparity point $(x,y)$, for each nearby point  $(x+x_{w},y+y_{w})$ within $\mathcal{N}(x,y)$, we estimate its \emph{consistency} with the central point as:

\vspace{-0.3cm}\begin{equation}
    W_{(x+x_{w},y+y_{w})} = e^{-\left(\frac{\left(x_{w}\right)^{2}+\left(y_{w}\right)^{2}}{2\sigma_{xy}^{2}}+\frac{\left(I\left(x+x_{w},y+y_{w}\right)-I\left(x,y\right)\right)^{2}}{2\sigma_{i}^{2}}\right)}
    \label{eq:adaptive_patch_weight}
\end{equation}
with $\sigma_{xy}$, $\sigma_{i}$ hyper-parameters. 
Then, we project a random value onto the two fictitious images only for those points with a similarity score $W_{(x+x_{w},y+y_{w})}$ higher than a threshold $t_\text{adpt}$, a hyper-parameter of the method. Fig. \ref{fig:adaptive} illustrates how the shape of the patch adapts to image content. Additionally, we update the upper threshold similarity scores in a data structure of the image size initialized to zero for each available sparse disparity point. Hence, we can project the random pattern with the highest score for overlapping patches. 

\subsubsection{Image Padding}

\begin{figure}[t]
    \centering
     \begin{tabular}{cc}
         \includegraphics[width=0.4\linewidth]{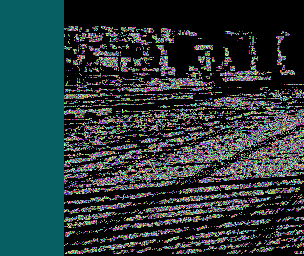} & \includegraphics[width=0.4\linewidth]{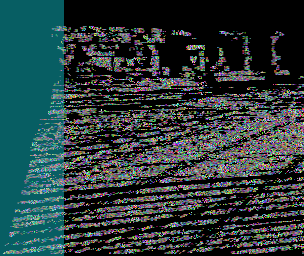} \\
     \end{tabular}
     \vspace{-0.32cm}
     \caption{\textbf{Left padding on a fictitious patterned stereo pair.} The padding is visually highlighted using a teal color on  $I_r$ (left) and $I_t$ (right). It allows us to project out-of-image warped points in $I_t$.}
     \label{fig:padding}
 \end{figure}

As for any stereo setup, our made of two virtual cameras inherits a well-known issue: the cameras do not frame a completely overlapping portion of the scene. Specifically, in our setup depicted in Fig. \ref{fig:overview}, the left border of the reference image will not be visible in the target image. However, since we have complete control over image generation, we can easily eliminate this issue by extending the field of view of our fictitious cameras on the left side, by applying image padding to account for the largest warped point out of the image $w_\text{out}$. Accordingly, we can project virtual patterns that otherwise would pop out the left image border. Ultimately, the only trick needed is left cropping the output dense disparity map. Fig. \ref{fig:padding} shows how padding works.

\section{Experiments}

We now present our experimental evaluation, including implementation details, datasets, and analysis of results.

\subsection{Implementation Details}

We implement RGB and Random pattern projection in Python, both with padding and adaptive patches. The hyper-parameters include $\sigma_{xy}=1$, $\sigma_{i}=1, t_\text{adpt}=0.001$ for adaptive patch.
We conduct experiments with various stereo matchers: RAFT-Stereo \cite{lipson2021raft}, IGEV-Stereo \cite{Xu_2023_CVPR}, GMStereo \cite{xu2022unifying}, PSMNet \cite{chang2018psmnet}, OpenCV's SGM implementation \cite{hirschmuller2007stereo, Menze2015CVPR}, and SDC \cite{choi2021stereodc}. We re-implemented SDC from scratch due to the unavailability of the source code. Depth refinement was omitted for a fairer comparison, although it has been shown to have a low impact on final accuracy \cite{yao2018mvsnet}. 
Then, we compare VPP4DC in its best setting against conventional completion frameworks: NLSPN \cite{park2020non}, CompletionFormer \cite{zhang2023completionformer}, and SpAgNet \cite{Conti_2023_WACV}.

We will consider several cross-domain settings, either from synthetic to real data or across real datasets. Purposely, we will set the SceneFlow dataset \cite{MIFDB16} as the synthetic training dataset: for stereo matchers, we use the authors' weights for IGEV-Stereo and GMStereo and retrained RAFT-Stereo and PSMNet following \cite{bartolomei2023ICCV}, while we train from scratch completion frameworks by adapting their recommended settings to this specific dataset.
Concerning generalization across real datasets, we perform either new training from scratch or fine-tuning of the synthetic pre-trained models.
A detailed description of the training protocols, loss functions, and any hyper-parameter setting is reported in the supplementary material.

\begin{table}[t]
\centering
\renewcommand{\tabcolsep}{6pt}
\scalebox{0.7}{
\begin{tabular}{|ccc|cc|}
\hline
\multicolumn{3}{|c|}{Hyperparameters} & \multicolumn{2}{|c|}{MAE (m)} \\
\hline
Patch size & Adaptive patch & Left Padding & NYU \cite{nyudepthv2} & KITTI DC \cite{kittidc} \\
\hline\hline

$1\times1$ & \xmark & \xmark & 0.090 & 0.412 \\
$1\times1$ & \xmark & \cmark & 0.094 & \underline{0.408} \\

$3\times3$ & \xmark & \xmark & 0.086 & 0.418 \\
$3\times3$ & \xmark & \cmark & 0.083 & 0.414 \\
$3\times3$ & \cmark & \xmark & 0.086 & 0.411 \\
$3\times3$ & \cmark & \cmark & 0.086 & \bf 0.406 \\

$5\times5$ & \xmark & \xmark & 0.084 & 0.431 \\
$5\times5$ & \xmark & \cmark & \underline{0.079} & 0.427 \\
$5\times5$ & \cmark & \xmark & 0.084 & 0.412 \\
\rowcolor{orange}
$5\times5$ & \cmark & \cmark & 0.080 & 0.409 \\

$7\times7$ & \xmark & \xmark & 0.083 & 0.485 \\
$7\times7$ & \xmark & \cmark & \bf 0.078 & 0.480 \\
$7\times7$ & \cmark & \xmark & 0.084 & 0.434 \\
$7\times7$ & \cmark & \cmark & 0.080 & 0.432 \\

$9\times9$ & \xmark & \xmark & 0.085 & 0.561 \\
$9\times9$ & \xmark & \cmark & \bf 0.078 & 0.556 \\
$9\times9$ & \cmark & \xmark & 0.082 & 0.486 \\
$9\times9$ & \cmark & \cmark & \bf 0.078 & 0.486 \\
\hline
\end{tabular}}
\vspace{-0.3cm}
\caption{
\textbf{Hyper-parameters study.}
Results on NYU and KITTI DC with a RAFT-Stereo model trained on synthetic data.
}\vspace{-0.3cm}
\label{tab:ablation_vpp}
\end{table}

\subsection{Evaluation Datasets \& Protocol}

Four datasets are used in our experiments, comprising scenes captured in both indoor and outdoor settings. 

\textbf{KITTI DC \cite{kittidc}.} The KITTI depth completion dataset depicts driving scenarios, captured at about 1280$\times$384 pixels resolution with sparse ground-truth depth maps collected using a LiDAR sensor. It provides 90K samples with RGB images, aligned sparse depth information, and semi-dense ground-truth data.
We use the original training/validation splits, 
and each sample is top-cropped by 100px and then center-cropped to $1216\times240$. 
Raw LiDAR points are filtered according to distance from the minimum \cite{aconti2022lidarconf} within a local $7\times7$ patch in any experiment.

\textbf{DDAD \cite{ddad}.} The DDAD dataset captures driving scenes using multiple synchronized cameras, measuring depths up to 250 meters. It includes about 12K training samples with RGB images and ground-truth depth maps aligned with each of the four cameras. The validation set comprises 3\,950 samples and ground-truth depth maps for each camera. We run experiments on validation samples from camera 1, where we sample about 20\% depth points from ground-truth -- filtered as done for KITTI raw LiDAR \cite{aconti2022lidarconf} -- and evaluate at 1936$\times$1216 resolution.

\textbf{NYU Depth V2 \cite{nyudepthv2}.} The NYU Depth V2 dataset comprises 464 indoor scenes captured using a Kinect sensor. Each sample was downsampled to $320\times240$ and center-cropped to $304\times228$.
As for the previous datasets, we used the original train/validation split for training and testing.

\textbf{VOID \cite{void}.} The VOID dataset provides synchronized color images at $640\times480$ and sparse depth maps of indoor and outdoor scenes. It includes about 1500 depth points obtained via XIVO \cite{feiWS19} together with dense ground-truth depth sensed by an active stereo camera. Of the total 56 sequences, 8 are used for testing purposes defining three benchmarks: VOID150 (150 points per frame), VOID500 (500 points), and VOID1500 (1500 points).

\textbf{Evaluation Protocol.} 
We assess the performance of all methods using root mean square error (RMSE) and mean absolute error (MAE).
Specifically, we evaluate our computed disparity maps converted into depth maps, for each pixel with valid ground-truth depth values. In the remainder, we will highlight the best results in a given configuration using \textbf{bold} font and indicate the absolute bests in \textbf{\textcolor{red}{red}}.

\begin{table}[t]
\centering
\renewcommand{\tabcolsep}{8pt}
\scalebox{0.6}{
\begin{tabular}{|l||rr|rr|rr|}
\multicolumn{1}{c}{ } & \multicolumn{6}{c}{MAE(m)} \\
\hline
 & \multicolumn{2}{c|}{RGB projection} & \multicolumn{2}{c|}{Synthesis Net \cite{choi2021stereodc}} & \multicolumn{2}{c|}{Random projection} \\ 
 Model & NYU & KITTI DC & NYU & KITTI DC & NYU & KITTI DC \\
 \hline\hline

 OpenCV-SGM \cite{hirschmuller2007stereo}  & 0.152 & 0.955 & 2.125 & 13.938 & 0.150 & 0.817 \\ 
\hline\hline
RAFT-Stereo \cite{lipson2021raft}         & \bf 0.096 & \bf 0.463 & 1.211 & 7.937 &\bf \textcolor{red}{0.080} &\bf \textcolor{red}{0.409} \\
IGEV-Stereo \cite{Xu_2023_CVPR}           & 0.119 & 0.609 & 0.908 & 6.178 & 0.107 & 0.604 \\ 
GMStereo \cite{xu2022unifying}            & 0.251 & 0.756 & 0.914 & 4.913 & 0.264 & 0.881 \\
PSMNet \cite{chang2018psmnet}             & 0.175 & 1.076 & 1.288 & 6.104 & 0.166 & 1.009  \\ 
\hline\hline 
SDC \cite{choi2021stereodc}  & 0.134 & 0.678 & \bf 0.293 & \bf 0.830 & 0.132 & 0.613 \\
\hline
\end{tabular}}
\vspace{-0.3cm}
\caption{\textbf{VPP4DC with off-the-shelf stereo networks.} Results on NYU and KITTI DC. Networks trained on SceneFlow \cite{MIFDB16}.\vspace{-0.3cm}
}
\label{tab:results_offshelf}
\end{table}

\subsection{Ablation Study}

We assess the impact of various design choices in VPP4DC, by validating over 100 images sampled from each of the KITTI DC, DDAD, NYU, and VOID training sets.

\textbf{Hyper-Parameters.} Tab. \ref{tab:ablation_vpp} collects the MAE achieved on the NYU and KITTI DC datasets by varying different hyper-parameters of the VPP4DC framework while maintaining the stereo model unchanged -- i.e., RAFT-Stereo \cite{lipson2021raft}. The virtual baseline is fixed at 0.15 and 0.54 for NYU and KITTI DC respectively.
We conduct experiments with the virtual pattern at pixel level or patches -- either without or with the adaptive pattern -- without or with left padding.
By applying padding to the pointwise pattern improves the results on KITTI DC, but leads to a small drop in accuracy on NYU. Switching to patchwise patterns yields consistent improvements on NYU, mainly due to the much sparser sets of depth points used on this benchmark compared to the raw LiDAR scans used in KITTI. However, it has a negative impact on KITTI, except when employing small -- $3\times3$ or $5\times5$ -- adaptive patches. With patches, left padding is generally beneficial on both datasets.
Based on this study, we can conclude that using a $7\times7$ adaptive patch yields the best results on NYU, while a $3\times3$ adaptive patch is optimal for KITTI DC, with padding being used for both.
Therefore, for the subsequent evaluations, we choose a configuration yielding a good trade-off on both datasets -- i.e., a $5\times5$ adaptive patch with left padding, in orange in the table.

\begin{figure}[t]
  \centering
  \includegraphics[trim=0cm 0.9cm 9cm 0cm, clip,width=0.45\textwidth]{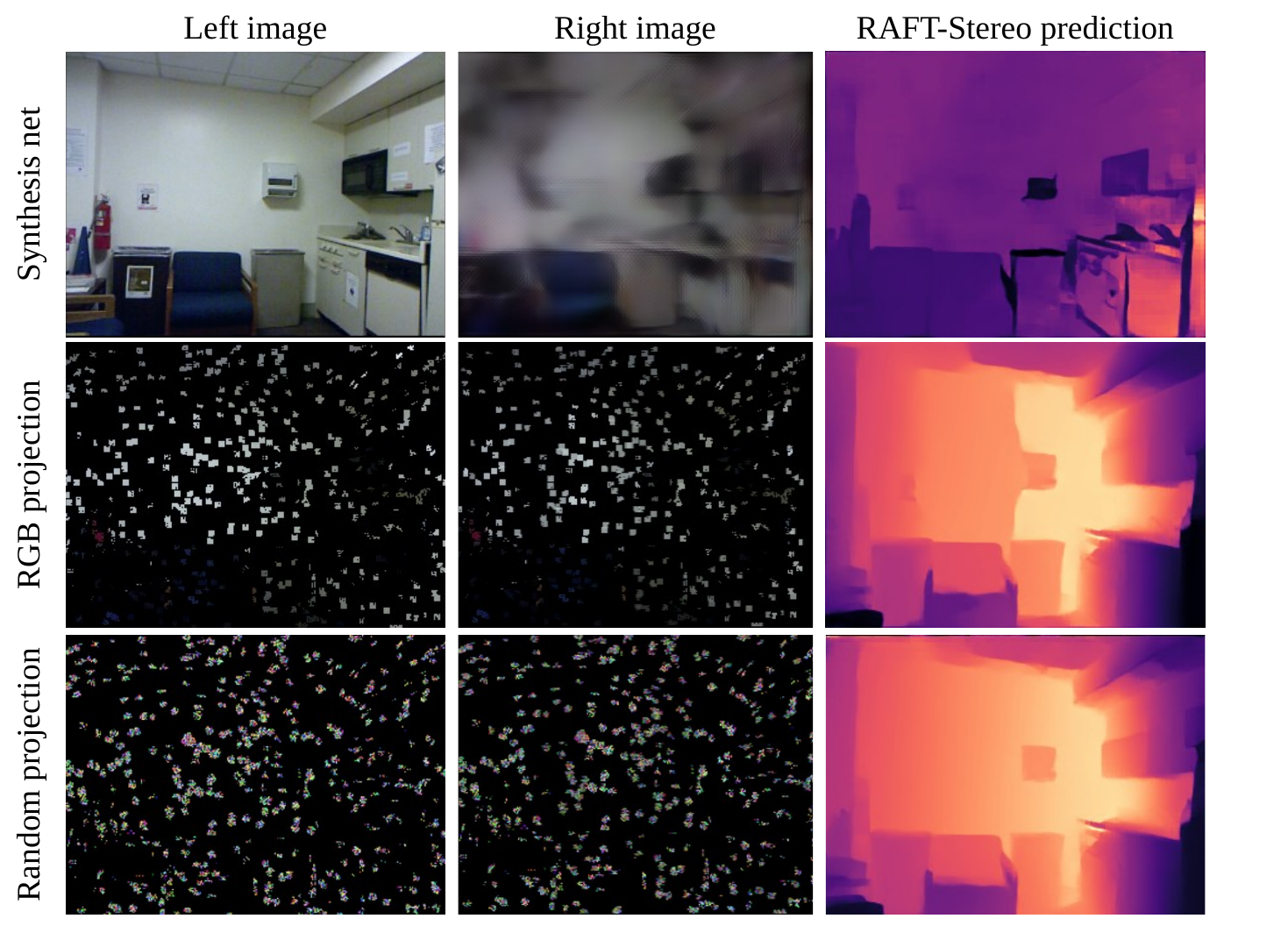}
  \vspace{-0.3cm}
  \caption{\textbf{Comparison of generated stereo pairs.} From top to bottom: stereo pairs by \cite{choi2021stereodc}, RGB pattern projection and Random pattern projection. For each, we report the output of RAFT-Stereo.}\vspace{-0.3cm}
  \label{fig:proj_methods}
\end{figure}

\begin{figure}[t]
  \centering
  \scalebox{1}{
  \renewcommand{\tabcolsep}{1pt}
  \begin{tabular}{ccc}
    & \scriptsize Right image (real) & \scriptsize Right image (synthesized) \\
    \rotatebox{90}{\hspace{0.5cm} \scriptsize SceneFlow} &\includegraphics[width=0.4\linewidth, trim=0 25 0 75, clip]{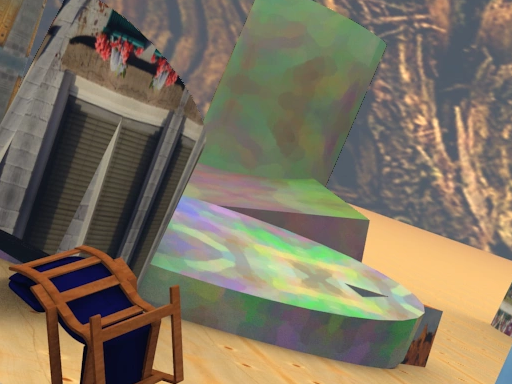} &
    \includegraphics[width=0.4\linewidth, trim=0 25 0 75, clip]{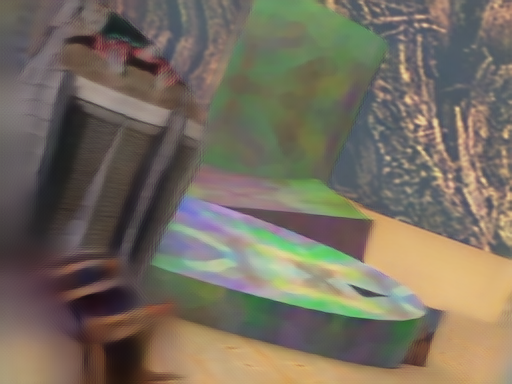} \\
    \rotatebox{90}{\hspace{0.1cm} \scriptsize KITTI DC} &\includegraphics[width=0.4\linewidth, trim=608 0 0 0, clip]{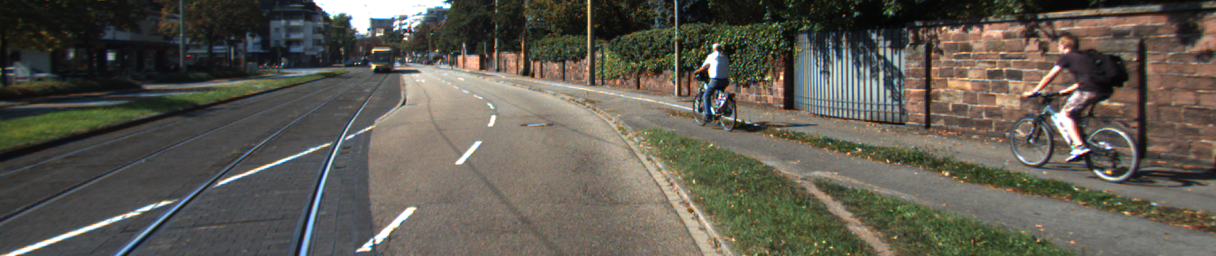} &
    \includegraphics[width=0.4\linewidth, trim=608 0 0 0, clip]{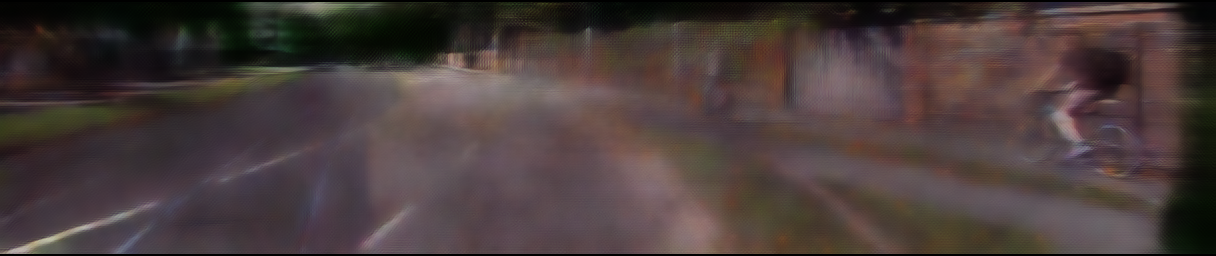} \\
  \end{tabular}
  }\vspace{-0.35cm}
  \caption{
    \textbf{Qualitative generalization analysis of synthesizer network \cite{choi2021stereodc}.}
    The network cannot handle high level details well.
  }\vspace{-0.35cm}
  \label{fig:synth_performance}
\end{figure}

\textbf{Stereo Models.} In Tab. \ref{tab:results_offshelf}, we evaluate VPP4DC in combination with various stereo matchers. The virtual baseline is fixed at 0.15 and 0.54 for NYU and KITTI DC respectively.
At the very top, we report the results obtained using the OpenCV SGM implementation \cite{hirschmuller2007stereo} as a reference, followed by recent stereo networks and SDC \cite{choi2021stereodc} trained on the SceneFlow dataset \cite{MIFDB16}.
On three different columns, we apply different strategies for generating the stereo images processed by the different methods: 5$\times$5 RGB projection, the use of an image synthesis model as proposed in \cite{choi2021stereodc}, and 5$\times$5 Random projection.
Starting from the left, we observe that PSMNet and GMStereo demonstrate lower accuracy compared to other networks and the SGM algorithm itself. We ascribe the better accuracy achieved by RAFT-Stereo and IGEV-Stereo to the use of a contextual network within their architecture. However, SDC compensates the lack of this latter by processing the sparse depth points as a direct input to the network itself.
By generating a dense right image through the synthesis network and processing it together with the real left image, the accuracy of any method drops, with SDC being the best-performing model under this setting. We attribute this outcome to the lack of high-frequency details in the synthetically generated right view due to the very low density of the depth points used to guide the synthesis process.
Finally, using the random pattern enables any model to achieve the lowest errors, with RAFT-Stereo proving to be the architecture better suited to exploit it. Thus, we use Random pattern projection coupled with RAFT-Stereo in our framework for the remaining experiments. We refer to it as VPP4DC.

\begin{figure}[t]
    \centering
    \includegraphics[width=0.9\linewidth]{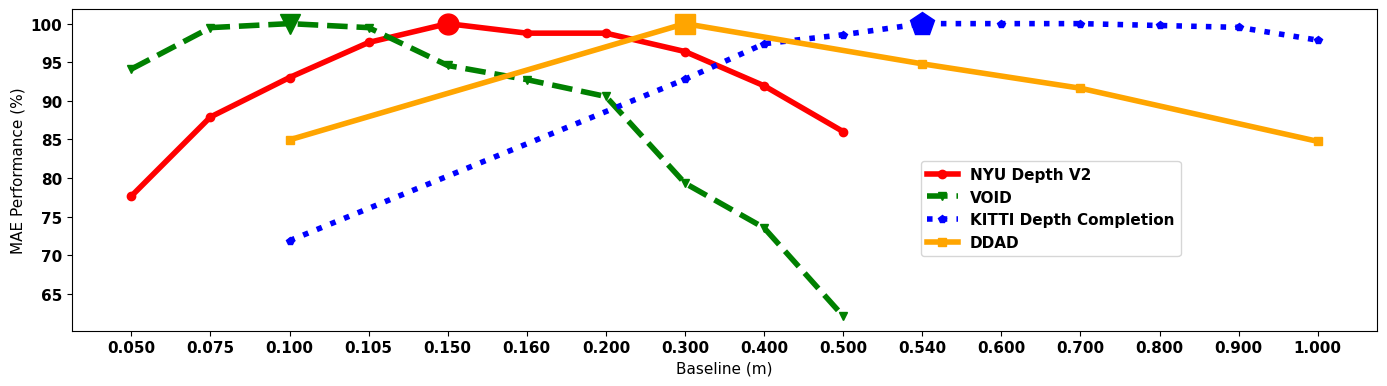}
    \vspace{-0.33cm}
    \caption{\textbf{Virtual baseline analysis.} VPP4DC performance with different virtual baselines.}\vspace{-0.3cm}
    \label{fig:baseline_tuning}
\end{figure}

\begin{table}[t]
\centering
\renewcommand{\tabcolsep}{8pt}
\scalebox{0.75}{
\begin{tabular}{|l|l|rr|}
\hline
Network & Test domain & RMSE (m) & MAE (m) \\
\hline\hline
NLSPN \cite{park2020non} & NYU \cite{nyudepthv2} & 0.716 & 0.440\\
SpAgNet \cite{Conti_2023_WACV} & NYU \cite{nyudepthv2} & 0.292 & 0.158 \\
CompletionFormer \cite{zhang2023completionformer}  & NYU \cite{nyudepthv2} & 0.374 & 0.186 \\
\hline VPP4DC (ours) &  NYU \cite{nyudepthv2} & \bf 0.247 & \bf 0.077 \\
\hline\hline
NLSPN \cite{park2020non} & VOID500 \cite{void} & 2.394 & 0.972 \\
SpAgNet \cite{Conti_2023_WACV} & VOID500 \cite{void} & 0.782 & 0.366 \\
CompletionFormer \cite{zhang2023completionformer} & VOID500 \cite{void} & 2.617 & 1.290 \\
\hline VPP4DC (ours) & VOID500 \cite{void} & \bf 0.614 & \bf 0.188 \\
\hline\hline
NLSPN \cite{park2020non} & KITTI DC \cite{kittidc} & 2.076 & 1.335 \\
SpAgNet \cite{Conti_2023_WACV} & KITTI DC \cite{kittidc} & 1.788 & 0.518 \\
CompletionFormer \cite{zhang2023completionformer} & KITTI DC \cite{kittidc} & 1.935 & 0.952 \\
\hline VPP4DC (ours) & KITTI DC \cite{kittidc} & \bf 1.609 & \bf 0.413 \\
\hline\hline
NLSPN \cite{park2020non} & DDAD \cite{ddad} & 11.612 & 3.498 \\
SpAgNet \cite{Conti_2023_WACV} & DDAD \cite{ddad} & 13.236 & 4.578 \\
CompletionFormer \cite{zhang2023completionformer} & DDAD \cite{ddad} & 9.959 & 2.518 \\
\hline VPP4DC (ours) & DDAD \cite{ddad} & \bf 7.303 & \bf 1.529 \\ 
\hline
\end{tabular}}
\vspace{-0.3cm}
\caption{
\textbf{Synthetic-to-real generalization.} All networks are trained on SceneFlow \cite{MIFDB16} and tested on real datasets.}\vspace{-0.3cm}
\label{tab:zeroshot_results}
\end{table}

\begin{table*}[t]
\centering
\renewcommand{\tabcolsep}{20pt}
\scalebox{0.6}{
\begin{tabular}{|l|rr|rr||rr|rr|}
 \multicolumn{1}{c}{ } & \multicolumn{4}{c}{Train on KITTI DC \cite{kittidc}} & \multicolumn{4}{c}{Train on NYU \cite{nyudepthv2}} \\
 \cline{2-9}
 \multicolumn{1}{c}{ } & \multicolumn{2}{|c|}{DDAD \cite{ddad}} & \multicolumn{2}{|c||}{VOID500 \cite{void}} & \multicolumn{2}{c|}{DDAD \cite{ddad}} & \multicolumn{2}{c|}{VOID500 \cite{void}} \\
 \hline
 Network & RMSE (m) & MAE (m) & RMSE (m) & MAE (m) & RMSE (m) & MAE (m) & RMSE (m) & MAE (m) \\
 \hline\hline
 NLSPN \cite{park2020non} &  11.646 & 4.621 & 5.627 & 4.196 & 20.180 & 6.882 & 0.802 & 0.381 \\
 SpAgNet \cite{Conti_2023_WACV} &  18.247 & 9.130 & 1.153 & 0.485 & 36.728 & 21.732 & \bf 0.752 & 0.326 \\
 CompletionFormer \cite{zhang2023completionformer} & \bf 9.606 &  3.328 & 11.640 & 9.856 & 16.479 & 5.649 & 0.821 & 0.385 \\
 \hline
 VPP4DC (ours) &  10.247 &  \bf 2.290 & \bf 0.934 & \bf 0.356 & \bf 9.246 & \bf 3.001 & 0.840 & \bf 0.307 \\
 \hline
 \multicolumn{9}{c}{\textbf{(a)}}\\

 \multicolumn{1}{c}{ } & \multicolumn{4}{c}{Train on SceneFlow \cite{MIFDB16} + KITTI DC \cite{kittidc}} & \multicolumn{4}{c}{Train on SceneFlow \cite{MIFDB16} + NYU \cite{nyudepthv2}} \\
 \cline{2-9}
 \multicolumn{1}{c}{ } & \multicolumn{2}{|c|}{DDAD \cite{ddad}} & \multicolumn{2}{|c||}{VOID500 \cite{void}} & \multicolumn{2}{c|}{DDAD \cite{ddad}} & \multicolumn{2}{c|}{VOID500 \cite{void}} \\
 \hline
 Network & RMSE (m) & MAE (m) & RMSE (m) & MAE (m) & RMSE (m) & MAE (m) & RMSE (m) & MAE (m) \\
 \hline\hline
 NLSPN \cite{park2020non} &  9.231 & 2.498 & 2.426 & 0.886 & 40.221 & 18.487 & 0.783 & 0.301 \\
 SpAgNet \cite{Conti_2023_WACV} &  17.540 & 9.202 & 0.878 & 0.458 & 36.878 & 21.808 & 0.765 & 0.342 \\
 CompletionFormer \cite{zhang2023completionformer} &  9.471 &  3.607 & 3.418 & 2.294 & 35.590 &  18.928 & 0.929 & 0.429 \\
 \hline
 VPP4DC (ours) & \bf  \textcolor{red}{7.048} & \bf \textcolor{red}{1.580} & \bf \textcolor{red}{0.582} & \bf \textcolor{red}{0.187} & \bf \textcolor{red}{6.781} &  \bf \textcolor{red}{1.344} & \bf \textcolor{red}{0.652} & \bf \textcolor{red}{0.187} \\
 \hline
 \multicolumn{9}{c}{\textbf{(b)}}\\
\end{tabular}}
\vspace{-0.4cm}
\caption{
\textbf{Real-to-real generalization.} Results in different train/test scenarios, without (a) and with (b) pre-training on SceneFlow \cite{MIFDB16}.
}\vspace{-0.35cm}
\label{tab:cross_results}
\end{table*}

\begin{table}[t]
\centering
\renewcommand{\tabcolsep}{5pt}
\scalebox{0.56}{
\begin{tabular}{|l|rr|rr|rr|}
 \multicolumn{1}{c}{ } & \multicolumn{6}{c}{Train on SceneFlow} \\
 \cline{2-7}
 \multicolumn{1}{c}{ } & \multicolumn{2}{|c|}{VOID150 \cite{void}} & \multicolumn{2}{|c|}{VOID500 \cite{void}} & \multicolumn{2}{|c|}{VOID1500 \cite{void}} \\
 \hline
 Network & RMSE (m) & MAE (m) & RMSE (m) & MAE (m) & RMSE (m) & MAE (m) \\
 \hline\hline
 NLSPN \cite{park2020non} & 4.989 & 2.523 & 2.394 & 0.972 & 1.353 & 0.427 \\
 SpAgNet \cite{Conti_2023_WACV} & 0.901 & 0.452 & 0.782 & 0.366 & 0.729 & 0.270 \\
 CompletionFormer \cite{zhang2023completionformer} & 4.131 & 2.369 & 2.617 & 1.290 & 1.684 & 0.642 \\
 \hline
 VPP4DC (ours) & \bf 0.748 & \bf 0.245 & \bf 0.614 & \bf 0.188 & \bf 0.606 & \bf 0.166 \\
 \hline
\end{tabular}}
\vspace{-0.3cm}
\caption{
\textbf{Density generalization.} Results in different train/test scenarios, without (a) and with (b) pre-training on SceneFlow \cite{MIFDB16}.
}\vspace{-0.5cm}
\label{tab:density_results_synth}
\end{table}

Fig. \ref{fig:proj_methods} shows a qualitative example of the results obtained by RAFT-Stereo using virtual stereo pairs generated through the three approaches. Using the left and synthesized right images yields poor results due to the lack of details in the generated view. We ascribe this behaviour to the lack of generalization capabilities by the synthesis network: Fig. \ref{fig:synth_performance} shows how images generated in the training domain appear detailed, with their quality degrading when applied to the KITTI DC dataset.
On the contrary, Fig. \ref{fig:proj_methods} shows that RGB projection produces qualitatively good results already, with Random projection further facilitating the recovery of fine structures in the final prediction.

\textbf{Virtual Baseline.} Fig. \ref{fig:baseline_tuning} plots how the virtual baseline length impacts the accuracy of VPP4DC. For each baseline $b$, we plot the relative accuracy as the minimum MAE achieved across all the baselines over the MAE computed with baseline $b$. 
In general, a tiny baseline dampens the resolution at farther distances, reducing accuracy in those regions. In contrast, a wide baseline produces larger disparity values, potentially falling outside the disparity distribution observed during training by most stereo networks \cite{MIFDB16}. We can observe similar trends across driving datasets, such as DDAD and KITTI DC, where the optimal baseline is 0.30 and 0.50m, respectively, with larger baselines yielding similar results for KITTI -- up to 0.80m -- while increasing the error on DDAD. We ascribe this to the different distribution of depth values observed in the two datasets, showing a higher percentage of pixels at closer distances on DDAD.
On datasets featuring a much lower depth range, such as NYU and VOID, the best results are achieved with shorter baselines, specifically 0.15m and 0.10m, respectively. From now on, we will use these optimal virtual baselines for the corresponding datasets.

\subsection{Synthetic-to-Real Generalization}

We now assess the robustness of state-of-the-art depth completion architectures and our VPP4DC framework to harsh domain shifts, starting from a synthetic-to-real transfer setup.
Specifically, we train VPP4DC and completion models \cite{park2020non,Conti_2023_WACV,zhang2023completionformer} on the SceneFlow dataset and evaluate their accuracy on the four real benchmarks -- NYU, VOID500, KITTI DC, and DDAD.
Three are the main challenges for these approaches: i) the highly diverse image content depicted in the various datasets, ii) the significantly different distributions of sparse depth points, and iii) the varying image resolutions, in particular on DDAD. 

Tab. \ref{tab:zeroshot_results} collects the outcome of this experiment. NLSPN often suffers the most from the domain shift, followed by CompletionFormer, partially compensating for it thanks to its global receptive field. SpAgNet proves to be highly robust by its design, specifically suited for handling different input distributions. Nevertheless, it faces significant challenges in adapting to DDAD, indicating limitations in handling high resolution. 
Finally, VPP4DC consistently achieves the highest accuracy,
making it the preferred option for addressing synthetic-to-real generalization for depth completion.

\begin{table*}[t]
\centering
\renewcommand{\tabcolsep}{7pt}
\scalebox{0.625}{
\begin{tabular}{|l|rr|rr|rr||rr|rr|rr|}
 \multicolumn{1}{c}{ } & \multicolumn{6}{c}{Train on KITTI DC \cite{kittidc}} & \multicolumn{6}{c}{Train on NYU \cite{nyudepthv2}} \\
 \cline{2-13}
 \multicolumn{1}{c}{ } & \multicolumn{2}{|c|}{VOID150 \cite{void}} & \multicolumn{2}{c|}{VOID500 \cite{void}} & \multicolumn{2}{c||}{VOID1500 \cite{void}} & \multicolumn{2}{c|}{VOID150 \cite{void}} & \multicolumn{2}{c|}{VOID500 \cite{void}} & \multicolumn{2}{c|}{VOID1500 \cite{void}}\\
 \hline
 Network & RMSE (m) & MAE (m) & RMSE (m) & MAE (m) & RMSE (m) & MAE (m) & RMSE (m) & MAE (m) & RMSE (m) & MAE (m) & RMSE (m) & MAE (m) \\
 
 \hline\hline
 
 NLSPN \cite{park2020non} & 6.587 & 4.933 & 5.627 & 4.196 & 4.784 & 3.591 & 0.963 & 0.492 & 0.802 & 0.381 & 0.737 & 0.298 \\
 SpAgNet \cite{Conti_2023_WACV} &  1.240 &  0.533 & 1.154 & 0.485 & 1.108 & 0.446 & \bf 0.866 & 0.408 & \bf 0.752 & 0.326 & \bf 0.706 & \bf 0.244 \\
 CompletionFormer \cite{zhang2023completionformer} & 15.543 & 13.490 & 11.640 & 9.856 & 8.804 & 7.446 & 0.956 & 0.487 & 0.821 & 0.385 & 0.726 & 0.261 \\
 \hline
 VPP4DC (ours) & \bf 1.118 & \bf 0.507 & \bf 0.934 & \bf 0.356 & \bf 0.789 & \bf 0.244 & 0.960 & \bf 0.397 & 0.840 & \bf 0.307 & 0.800 & 0.253 \\
 \hline
 \multicolumn{13}{c}{\textbf{(a)}} \\

 \multicolumn{1}{c}{ } & \multicolumn{6}{c}{Train on SceneFlow \cite{MIFDB16} + KITTI DC \cite{kittidc}} & \multicolumn{6}{c}{Train on SceneFlow \cite{MIFDB16} + NYU \cite{nyudepthv2}} \\
 \cline{2-13}
 \multicolumn{1}{c}{ } & \multicolumn{2}{|c|}{VOID150 \cite{void}} & \multicolumn{2}{c|}{VOID500 \cite{void}} & \multicolumn{2}{c||}{VOID1500 \cite{void}} & \multicolumn{2}{c|}{VOID150 \cite{void}} & \multicolumn{2}{c|}{VOID500 \cite{void}} & \multicolumn{2}{c|}{VOID1500 \cite{void}}\\
 \hline
 Network & RMSE (m) & MAE (m) & RMSE (m) & MAE (m) & RMSE (m) & MAE (m) & RMSE (m) & MAE (m) & RMSE (m) & MAE (m) & RMSE (m) & MAE (m) \\
 
 \hline\hline
 
 NLSPN \cite{park2020non} & 4.616 & 1.991 & 2.426 & 0.886  & 1.349 & 0.448 & 1.138 & 0.535 & 0.783 & 0.301 & 0.668 & 0.210 \\
 SpAgNet \cite{Conti_2023_WACV} & 0.959 & 0.503 & 0.878 & 0.458 & 0.875 & 0.420 & 0.874 & 0.418 & 0.766 & 0.342 & 0.718 & 0.253 \\
 CompletionFormer \cite{zhang2023completionformer} & 5.396 & 3.736 & 3.418 & 2.294 & 2.446 & 1.608 & 1.162 & 0.628 & 0.929 & 0.429 & 0.800 & 0.287 \\ 
 \hline
 VPP4DC (ours) & \bf \textcolor{red}{0.690} & \bf \textcolor{red}{0.247} & \bf \textcolor{red}{0.582} & \bf \textcolor{red}{0.187} & \bf \textcolor{red}{0.543} & \bf \textcolor{red}{0.148} & \bf \textcolor{red}{0.865} & \bf \textcolor{red}{0.259} & \bf \textcolor{red}{0.652} & \bf \textcolor{red}{0.187} & \bf \textcolor{red}{0.595} & \bf \textcolor{red}{0.157} \\
 \hline
 \multicolumn{13}{c}{\textbf{(b)}} \\

\end{tabular}}
\vspace{-0.4cm}
\caption{
\textbf{Density generalization.} Results in different train/test scenarios, without (a) and with (b) pre-training on SceneFlow \cite{MIFDB16}.
}\vspace{-0.5cm}
\label{tab:density_results_real}
\end{table*}

\begin{table}[t]
\centering
\renewcommand{\tabcolsep}{5pt}
\scalebox{0.75}{
\begin{tabular}{|l|rr|rr|}
\multicolumn{1}{c}{ } & \multicolumn{2}{c}{NYU} & \multicolumn{2}{c}{KITTI DC} \\
\hline
Network & RMSE (m) & MAE (m) & RMSE (m) & MAE (m)\\
\hline\hline
SCPU \cite{scpu} & 0.544 & 0.240 & 1.591 & 0.329 \\

\hline
NLSPN \cite{park2020non} & 0.092 & 0.035 & \bf \textcolor{red}{0.772} & \bf \textcolor{red}{0.197} \\
SpAgNet \cite{Conti_2023_WACV} & 0.114 & 0.045 & 0.856 & 0.222 \\
CompletionFormer \cite{zhang2023completionformer} & \bf \textcolor{red}{0.090} & \bf \textcolor{red}{0.035} & 0.849 & 0.216 \\
\hline
VPP4DC (ours) & 0.117 & 0.044 & 0.999 & 0.269 \\ 

\hline
\multicolumn{5}{c}{\textbf{(a)}} \\

\multicolumn{1}{c}{ } & \multicolumn{2}{c}{NYU} & \multicolumn{2}{c}{KITTI DC} \\
\hline
Network & RMSE (m) & MAE (m) & RMSE (m) & MAE (m)\\
\hline\hline
SCPU \cite{scpu} & 0.544 & 0.240 & 1.591 & 0.329 \\

\hline
NLSPN \cite{park2020non} & 0.123 & 0.051 & 1.129 & 0.353 \\
SpAgNet \cite{Conti_2023_WACV} & 0.104 & 0.043 & \bf 0.832 & \bf 0.216 \\
CompletionFormer \cite{zhang2023completionformer} & \bf 0.102 & \bf 0.042 & 1.090 & 0.259 \\
\hline
VPP4DC (ours) & 0.119 & 0.048 & 1.095 & 0.304 \\

\hline
\multicolumn{5}{c}{\textbf{(b)}} \\
\end{tabular}}
\vspace{-0.4cm}
\caption{
\textbf{In-domain performance.} Results without (a) and with (b) pre-training on SceneFlow \cite{MIFDB16}.
}\vspace{-0.3cm}
\label{tab:indomain}
\end{table}

\subsection{Real-to-Real Generalization}

The synthetic-to-real experiment already highlights the potentially superior robustness of VPP4DC compared to well-established completion frameworks across different datasets.
To confirm this, we conduct further experiments by training each method on real datasets and evaluating their accuracy in unseen, different environments.
Purposely, we choose NYU and KITTI DC as the training datasets, as they have historically been the most commonly used in the depth completion literature. We then test the models on VOID500 and DDAD, creating the KITTI-to-DDAD and NYU-to-VOID500 benchmarks, where the training and testing domains are quite similar. Additionally, we establish the KITTI-to-VOID500 and NYU-to-DDAD ones, where the domain shift is more severe.
Tab. \ref{tab:cross_results} summarizes the results of this experiment, conducted under two settings.
 
On top (a), where the models are trained on either KITTI DC or NYU only, we observe that NLSPN and CompletionFormer underperform when trained on datasets significantly different from the testing domain, i.e., in the KITTI-to-VOID500 setup or the NYU-to-DDAD. SpAgNet performs the worst on DDAD while it shows considerable robustness in the KITTI-to-DDAD setting and achieves the absolute best RMSE on NYU-to-VOID500. VPP4DC demonstrates strong generalization either when trained on KITTI DC or NYUv2 solely, yielding the lowest error in most cases. 

At the bottom (b), we repeat the same evaluation, this time using models being pre-trained on SceneFlow -- firstly, we pre-train the models on SceneFlow, where the completion networks handle the completion task and VPP4DC focuses on stereo matching. Then, we fine-tune these pre-trained models on either KITTI DC or NYU for evaluation.
Overall, pre-training appears to be beneficial for models that are subsequently trained on KITTI DC, as it improves their accuracy in most cases. Conversely, it has a negative impact -- negligible or not -- when the models are eventually trained on NYU, except for VPP4DC.
Indeed, our method consistently achieves the best results when empowered by stereo pre-training -- which is crucial for properly learning how to match images, either real or generated through our pattern -- largely reducing RMSE and MAE on VOID500 especially.

\subsection{Density Generalization}

We investigate model robustness to varying input depth point densities across domains using the VOID150, VOID500, and VOID1500 benchmarks.

\textbf{Synthetic-to-Real.} We start by evaluating networks trained on the SceneFlow dataset \cite{MIFDB16}. Tab. \ref{tab:density_results_synth} summarizes the outcome of this experiment. Among the established completion frameworks, SpAgNet demonstrates outstanding robustness to the different densities, while NLSPN and CompletionFormer suffer about $2\times$ increase of RMSE and MAE from VOID1500 to VOID500 and from this latter to VOID150.
Once again, VPP4DC surpasses all other models, affirming its robustness in this aspect too.

\textbf{Real-to-Real.} Next, we study the impact of density when the networks have been trained on KITTI DC or NYU datasets. Tab. \ref{tab:density_results_real} collects the results on the three VOID benchmarks. 
On top (a), we report the results for the models trained directly on real data. SpAgNet confirms its robustness to the varying input density. Moreover, when trained on a similar domain, such as the one of the NYU dataset, it often achieves the best results. It also outperforms VPP4DC, which proves to be more effective when generalizing from KITTI DC to VOID.
At the bottom (b), we evaluate models pre-trained on SceneFlow. We observe how this pre-training strategy improves the results when moving from KITTI DC to VOID splits -- i.e., very different domains -- while it has a negligible or negative impact when moving from NYU to VOID -- i.e., more similar ones.
On the contrary, VPP4DC consistently benefits from stereo pre-training on SceneFlow, often halving the errors and emerging as the most robust framework to varying input density.

\subsection{In-Domain Performance}

We conclude by evaluating the accuracy achieved by VPP4DC when training and testing on the same domains. Tab. \ref{tab:indomain} reports the outcome of this final experiment on two sub-tables. On top of both, we include the results achieved by a hand-crafted completion solution \cite{scpu} as a reference.

On top (a), each model is trained directly on the training set of the target dataset -- NYU or KITTI DC. Notably, the well-established completion frameworks outperform VPP4DC concerning specialization within a single domain.
At the bottom (b), we show the performance obtained by pre-training the networks on SceneFlow. 
Except for SpAgNet, the pre-training impacts negatively on the accuracy over the very same domain. Indeed, the absolute best results are obtained by CompletionFormer and NLSPN, on NYU and KITTI DC, respectively, when trained only on the target domain training set.

Despite being slightly less effective at specializing in single domains, VPP4DC still achieves competitive results, not far from the state-of-the-art. Nonetheless, we believe this small gap is a moderate price to pay for achieving much stronger generalization across very different domains. This makes VPP4DC better suited for in-the-wild deployment and a first significant step towards studying depth completion solutions with such capabilities.

\section{Conclusion}

This paper proposed a novel paradigm to achieve depth completion robust against the training environment.
Combining a state-of-the-art stereo network and our framework achieves competitive results on the same domain evaluation and greatly outperforms other task-specific networks in cross-domain generalization.
We believe this can pave the way for deploying depth completion in countless and exciting practical application contexts.

{
    \small
    \bibliographystyle{ieeenat_fullname}
    \bibliography{main}
}

\newpage\phantom{Supplementary}
\multido{\i=1+1}{11}{


\section*{Supplementary material}

# Revisiting Depth Completion from a Stereo Matching Perspective for Cross-domain Generalization – supplementary material

Luca Bartolomei[*,†]  Matteo Poggi[*,†]  Andrea Conti[†]  Fabio Tosi[†]  Stefano Mattoccia[*,†]

[*]Advanced Research Center on Electronic System (ARCES)
[†]Department of Computer Science and Engineering (DISI)
University of Bologna, Italy
{luca.bartolomei5, m.poggi, andrea.conti35, fabio.tosi5 stefano.mattoccia}@unibo.it
https://vppdc.github.io/

This document provides additional details regarding the 3DV 2024 paper "Revisiting Depth Completion from a Stereo Matching Perspective for Cross-domain Generalization".

## 1. Implementation details

In this section, we report additional details concerning our experiments.

### 1.1. Stereo matching networks

We now describe the training protocol adopted for stereo architectures involved in our experiments. In this phase, none of the models sees any virtual patterned image – i.e., they are trained only on the original, vanilla stereo pairs.

- **RAFT-Stereo [8].** We trained this stereo network using Sceneflow [9] synthetic dataset, following the original training protocol.

- **PSMNet [1].** We trained PSMNet using Sceneflow [9], following the original training protocol.

- **GMStereo [18], IGEV-Stereo [17].** We use authors' weights to test the networks with our framework.

- **SDC [2] – stereo module (CCVNorm [15]).** We trained CCVNorm on [9] using the PSMNet training protocol.

- **SDC [2] – view synthesis module.** Following the authors' guidelines, we employed a U-Net [13] style network with a double encoder and a single decoder. The first encoder processes context RGB image and produces multi-scale features $\mathcal{F}^s_{\text{rgb}}$. The second encoder processes sparse warped RGB images and produces multi-scale features $\mathcal{F}^s_{\text{warp}}$. Then, $\mathcal{F}^s_{\text{rgb}}$ and $\mathcal{F}^s_{\text{warp}}$ are concatenated at each scale in the channel dimension and fed to the decoder that generates a dense synthetic right image. We trained the image synthesizer network for 15 epochs on synthetic data [9], using a batch size of 2 and a learning rate of 0.0001. We used AdamW as the optimizer and applied a OneCycle schedule. We generate sparse points randomly sampling 1% of groundtruth data. We also employ RGB augmentation (i.e., color jittering, channels shuffling, histogram equalization) and spatial augmentation (i.e., horizontal flip, random $512 \times 384$ crop). Following [2], we employed the $\mathcal{L}_{\text{sdc}}$ loss:

$$\mathcal{L}_{\text{sdc}} = \mathcal{L}_{\text{photo}} + \mathcal{L}_{\text{vgg}} \quad (1)$$

$\mathcal{L}_{\text{sdc}}$ loss is a sum of two terms: $\mathcal{L}_{\text{photo}}$ (Eq. 2) regularizes the synthetic image $I_s$ with the groundtruth real image $\hat{I}$, while $\mathcal{L}_{\text{vgg}}$ (Eq. 3) indirectly regularizes $I_s$ by enforcing its features representation $\mathcal{F}_s$ to be close to features representation $\hat{\mathcal{F}}$ of $\hat{I}$. Features $\mathcal{F}_s$ and $\hat{\mathcal{F}}$ are computed using a pretrained VGG.

$$\mathcal{L}_{\text{photo}} = \mathbb{E}\left[\lambda \frac{1 - \text{SSIM}(I_s, \hat{I})}{2} + (1-\lambda)\left|I_s - \hat{I}\right|\right] \quad (2)$$

Where $\lambda = 0.85$ is a hyperparameter that weight structure similarity loss and $\mathcal{L}_1$ loss.

$$\mathcal{L}_{\text{vgg}} = \mathbb{E}\left[\left(\mathcal{F}_s - \hat{\mathcal{F}}\right)^2\right] \quad (3)$$

Differently from SDC's protocol, we employed a $\mathcal{L}_2$ loss instead of KL-divergence loss for $\mathcal{L}_{\text{vgg}}$ term since we encountered numerical stability issues using original loss.
- **OpenCV SGM [6].** To conclude, we report the parameters used for OpenCV SGM algorithm. We set P1, P2, BlockSize to 1176, 4704 and 16 respectively. We apply background interpolation [10] to fill invalid disparity values.

### 1.2. Depth completion networks

We now describe in detail the training protocol adopted for the conventional completion networks used in our evaluation.
- **CompletionFormer [19].** At first, using the authors' training protocol, we train the network using synthetic data [9]. Specifically, we train in depth domain for 20 epochs, sampling 1% random points from the groundtruth to generate sparse measurements. The learning rate was set to $0.001$ and batch size to 4. We set a depth cap to 90 meters for numerical stability. In addition to the authors' proposed data augmentations, we used $512 \times 384$ random crop. Starting from a synthetic checkpoint, we also fine-tuned the network on real indoor [11] and outdoor [14] datasets. Specifically, on NYU [11] we trained CompletionFormer for 20 epochs with batch size 16, a learning rate set to $0.0001$ and a depth clamp at 10 meters. Five hundred depth seeds were sampled randomly from groundtruth. We trained on full image size, discarding random crops. Instead, on KITTI DC [14] we trained for 10 epochs with batch size 4, a learning rate set to $0.0001$ and a depth clamp at 90 meters. We used the raw LiDAR data as the depth seed source. We trained on full image size, discarding random crops. Finally, for training on KITTI DC and NYU from scratch, we used the original authors' weights.
- **NLSPN [12] and SpAgNet [4].** Starting from the authors' proposed training protocols, we train the network on synthetic data [9] using Adam for 20 epochs at batch size 6 and a learning rate set to $0.001$. We generated deep seeds randomly sampling 1% of groundtruth data. As in [19], we set a depth cap at 90 meters, and used random $512 \times 384$ crop in addition to the original data augmentation protocol. From synthetic pre-training, we fine tuned in real indoor [11] and outdoor [14] scenarios. On NYU [11], we trained those networks for 20 epochs at batch size 20, a learning rate set to $0.0001$ and a depth cap at 10 meters. We sampled 500 depth seed from groundtruth as sparse measurements. We trained on full image size, discarding random crops. Instead, on KITTI DC [14] we apply the same parameters described before for CompletionFormer. Finally, for training on KITTI DC and NYU from scratch, we used original authors' weights. When evaluating with NLSPN [12] on DDAD [5], we downsampled at half resolution due to memory constraints, then after upsampling the predicted depth map, we evaluated at full resolution.
- **VPP4DC.** As mentioned in the main paper, we selected RAFT-Stereo [8] as the reference stereo network in our framework. Starting from the original training protocol, we trained on KITTI DC [14] from scratch for 10 epochs with batch size 4 and a learning rate of $0.0001$. As done before, we applied a depth clamp at 90 meters. To save memory and time, we used $608 \times 240$ random crop in addition to spatial augmentation (i.e., horizontal flip) and RGB augmentation (i.e., color jittering, channels shuffling, histogram equalization). On the same dataset, we fine-tuned from Sceneflow [9] pre-train using the same configuration aforementioned, except for the learning rate set to $0.00001$. Instead, on NYU [11], we trained from scratch for 20 epochs at batch size 12, a learning rate of $0.0001$ and a depth clamp at 10 meters. Five hundred points were randomly sampled from groundtruth as depth seeds. We trained on full image size, discarding random crops. On the same dataset, we fine-tuned from Sceneflow [9] pre-train using the same configuration aforementioned, except for the learning rate set to $0.00001$.

All depth completion training and fine tuning use a $\mathcal{L}_{12}$ loss:

$$\mathcal{L}_{12} = \mathbb{E}\left[|\mathbf{z}_r - \hat{\mathbf{z}}_r|\right] + \mathbb{E}\left[(\mathbf{z}_r - \hat{\mathbf{z}}_r)^2\right] \quad (4)$$

For traditional depth completion method SCPU [7], we set the hyper parameter of maximum depth equal to 10 meters for NYU [11] and VOID, keeping the original hyperparameters for KITTI DC [14]. Instead, for DDAD [5], we set maximum depth equal to 250 meters. Finally, for RAL [20] depth completion method, we disabled internal occlusion system as outdoors methods are already filtered using [3]. Also, we deploy small changes to execute the algorithm to other datasets rather than KITTI DC [14]: we removed depth clamps and hardcoded depth map shape.

| Network | Test domain | RMSE (m) | MAE (m) |
|---|---|---|---|
| SCPU [7] | NYU [11] | 0.544 | 0.240 |
| RAL [20] | NYU [11] | 0.495 | 0.237 |
| VPP4DC [6] (Ours) | NYU [11] | **0.321** | **0.129** |
| SCPU [7] | VOID500 [16] | 1.749 | 1.347 |
| RAL [20] | VOID500 [16] | 1.129 | 0.481 |
| VPP4DC [6] (Ours) | VOID500 [16] | **0.707** | **0.260** |
| SCPU [7] | KITTI DC [14] | 1.591 | **0.329** |
| RAL [20] | KITTI DC [14] | **1.571** | 0.345 |
| VPP4DC [6] (Ours) | KITTI DC [14] | 3.442 | 0.835 |
| SCPU [7] | DDAD [5] | 29.144 | 14.389 |
| RAL [20] | DDAD [5] | **7.383** | **1.579** |
| VPP4DC* [6] (Ours) | DDAD [5] | 8.763 | 2.254 |

Table I. **Capabilities of traditional algorithms in LiDAR-only depth completion.** The adaptive patch approach is turned off in VPP4DC to ignore the RGB context information. * With a baseline of 0.22 meters and maximum disparity of 256 pixels.

| | | RAFT-Stereo[8] | | IGEV-Stereo[17] | |
|---|---|---|---|---|---|
| | Test domain | RMSE (m) | MAE (m) | RMSE (m) | MAE (m) |
| With context | NYU [11] | **0.247** | **0.077** | 0.324 | **0.107** |
| Without context | NYU [11] | 0.343 | 0.151 | **0.277** | 0.132 |
| With context | KITTI DC [14] | **1.609** | **0.413** | 2.258 | **0.604** |
| Without context | KITTI DC [14] | 5.567 | 1.448 | 2.430 | 0.700 |

Table II. **Capabilities of context-aware stereo networks without context.** As expected, additional information from the RGB context image helps reduce errors.

## 2. Additional experiments

We investigate the impact of using the RGB contextual information on the performance of traditional stereo matchers such as SGM [6] and context-aware deep learning stereo matchers [8, 17].

**LiDAR-only depth completion using traditional stereo matching.** We now assess the performance of VPP4DC using a popular traditional stereo algorithm (i.e., OpenCV-SGM [6]) with respect to other handcrafted depth completion methods [7, 20]. Table I shows the accuracy achieved on four different indoor and outdoor datasets (i.e., NYU [11], VOID [16], KITTI DC [14], DDAD [5]). Not surprisingly, RAL and SCPU outperform VPP4DC with KITTI DC evaluation since these traditional algorithms are conceived ad-hoc for this dataset. However, moving to a similar outdoor dataset such as DDAD, SCPU accuracy drops compared to other methods: this evidence proves that SCPU is too specialized on the KITTI DC dataset. RAL is the absolute winner, reducing by 30% the MAE error compared to the second-best method VPP4DC. Instead, when facing challenging indoor datasets, VPP4DC outperforms all other traditional depth completion methods, showing a better capability to handle these different domains.

**Effects of context on context-aware stereo networks.** Given a context-aware stereo matching network such as RAFT-Stereo [8] or IGEV-Stereo [17], we ablate the impact of the RGB context feeding such networks with the real input image or a black image as context. Table II shows the result of this experiment. As expected, the RGB image context is highly beneficial for both networks – especially for RAFT-Stereo – and crucial for achieving the best results, as shown in Figure I.

## 3. Design choices: qualitative results

We show additional qualitative results about the proposed virtual projection method to demonstrate the effects of i) random projection; ii) raw depth occlusion filtering; iii) left padding; iv) adaptive patch.

- **(i) Uniform areas in RGB projection.** Figure II shows a qualitative view of a KITTI DC [14] fictitious scene projected using RGB projection (up) and random projection (bottom), respectively. RGB projection shows a large white uniform area due to the massive direct light on the camera sensor. This uniform area propagates in the fictitious stereo pair with RGB projection; consequently, the stereo pair is ambiguous. In contrast, Random projection does not rely on RGB information to generate a fictitious stereo pair, inherently making ambiguous uniform areas distinctive.
- **(ii) Artifacts on raw sparse depth map.** When facing depth completion setups where the RGB camera and depth sensors do not share the very same view, occlusions appear in the camera view's sparse depth map. As shown in the top of Figure III, when we fed this raw depth map to our virtual projection framework, repetitive patterns appeared: occlusions lead the

14

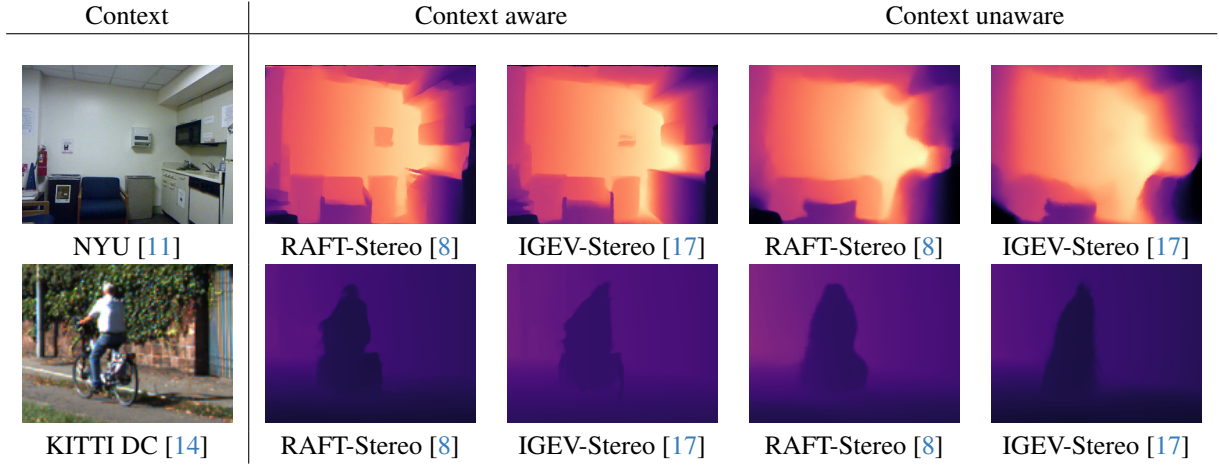

Figure I. **Visual effects of RGB context on stereo networks.** The RGB image allows the networks to recover better fine details not available in the sparse fictitious images generated by our projection method.

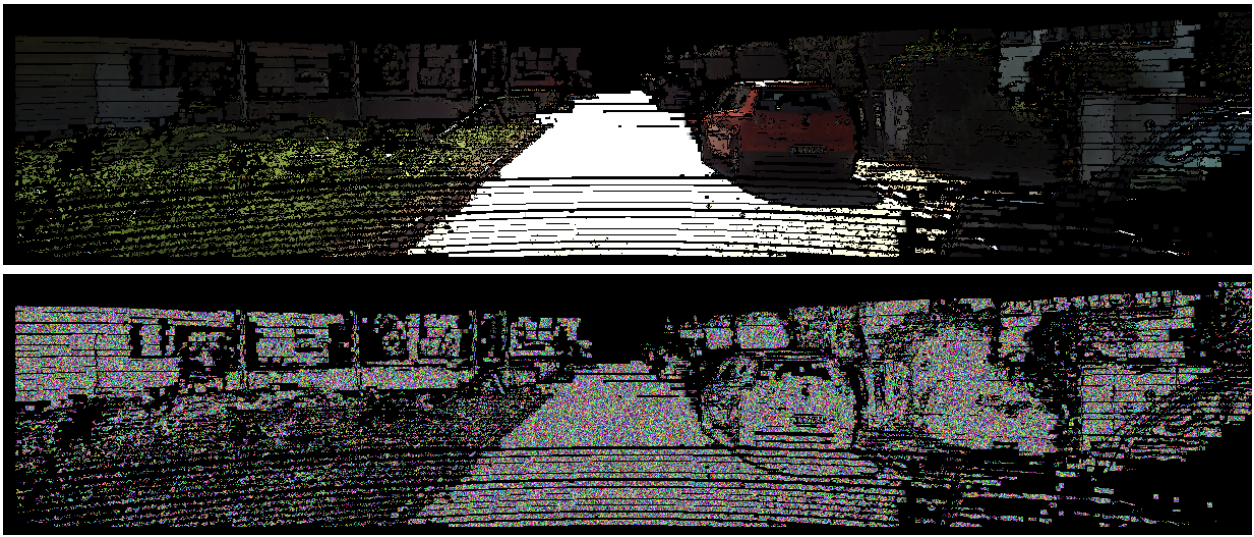

Figure II. **Qualitative view on uniform areas.** RGB projection keeps uniform regions as such, while random projection makes them distinctive.

stereo matching algorithm to match the same region (e.g., the right side of cyclist's body) to multiple locations. However, using the filtering technique [3] (Fig. III, bottom), the artifacts above vanish, and only the real matching region remains.
- **(iii) Recovering out-of-image depth seeds using left padding.** We qualitatively investigate the behaviour of RAFT-Stereo [8] and OpenCV-SGM[6] stereo matchers when facing left padding. Figure IV shows the effects of left padding on the stereo matchers above. In this particular indoor scenario [11], differences are less visible since there are only a few out-of-border depth seeds. We argue that these scarce depth points add a marginal contribution to the dense depth reconstruction process. Instead, in the outdoor scenario [14], we can appreciate a stronger recovery of out-of-border objects, such as the road blue sign.
- **(iv) Adaptive patch.** Figure V shows the effects of the adaptive patch used for an outdoor scenario [14]. Given a context image (Fig. V, left), our methodology shapes the original block-based patch (Fig. V, center) to a more appropriate silhouette (Fig. V, right). Supported by the figure above and hyper-parameters ablation, we argue that this method helps to preserve fine details that otherwise would have been lost.

15

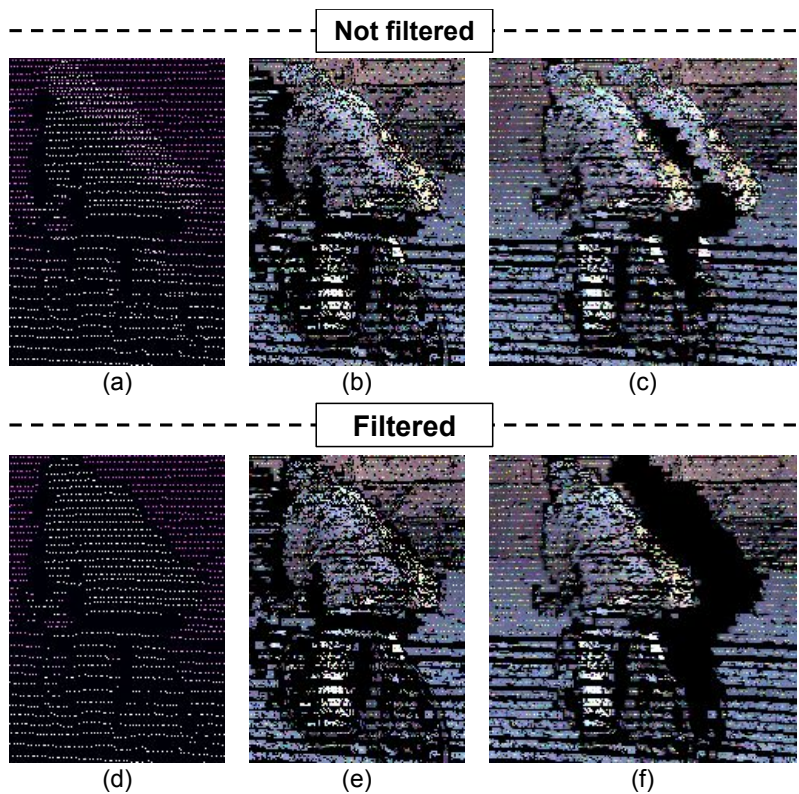

Figure III. **Qualitative effects of depth occlusion filtering.** The different position between the RGB camera and depth sensor (a) yields artifacts in virtual projection (b,c). Filtering out occluded points according to [3] (d) significantly softens the problem (e,f).

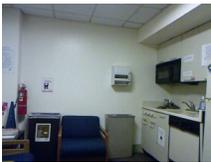

Figure IV. **Qualitative effects of left padding.** Effects are less visible in indoor dataset [11], while they are more apparent in outdoor scenarios [14]. Without left padding, stereo matchers struggle on left occluded objects (e.g., the blue road sign).

16

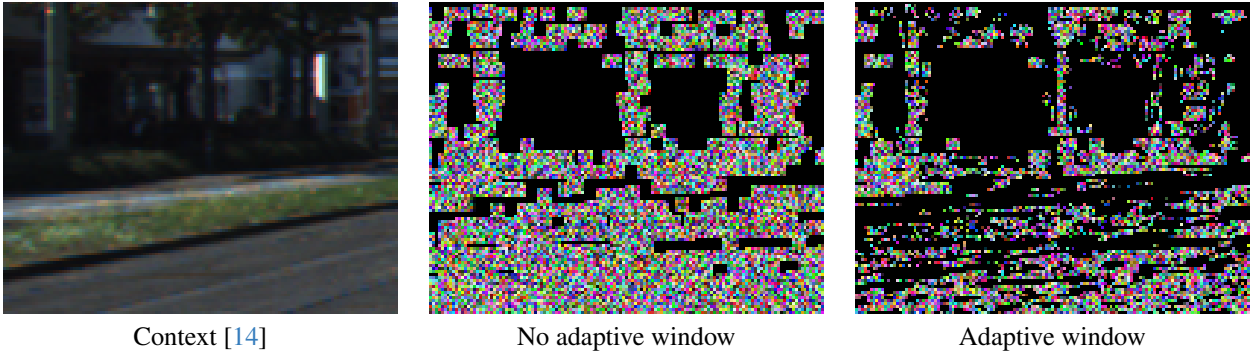

Context [14]        No adaptive window        Adaptive window

Figure V. **Qualitative effects of adaptive patch.** Given a context image, our proposal adapts the patch to objects in the scene (e.g., the tree at the center).

## 4. Virtual baseline study

In this section, we complement Fig. 7 in the main paper by adding a comparison between different virtual baselines in outdoor datasets [5, 14]. In particular, we inspect the discrepancy between the virtual baseline used in DDAD [5] and the one used in KITTIDC [14], – both acquired in a similar environment – in Fig. VI and VII. Firstly, we compare the depth distribution of the two datasets in Fig. VI (a) for the near depth range – *i.e.*, 0-10m. Supported by the distribution shown in the figure and the fact that both datasets are acquired in similar conditions, we agree that both outdoor datasets share similar close-range depth distributions. Then, we investigate the inverse close-depth function for both datasets in Fig. VI (b), using the focal length given by the real camera and the optimal virtual baseline given by Fig. 7. Additionally, we plot the inverse close-depth function for DDAD [5] with a larger virtual baseline, equal to 0.54m – *i.e.*, the same virtual baseline of KITTIDC [14]. As expected, the larger virtual baseline in [5] leads to a larger disparity range, even ∼3 times the maximum disparity registered for KITTIDC [14] and ∼2 times the maximum disparity registered using the optimal virtual baseline. This extended range might hinder a deep stereo model not used to deal with it, especially in the close-depth regions, such as the road surface just under the camera.

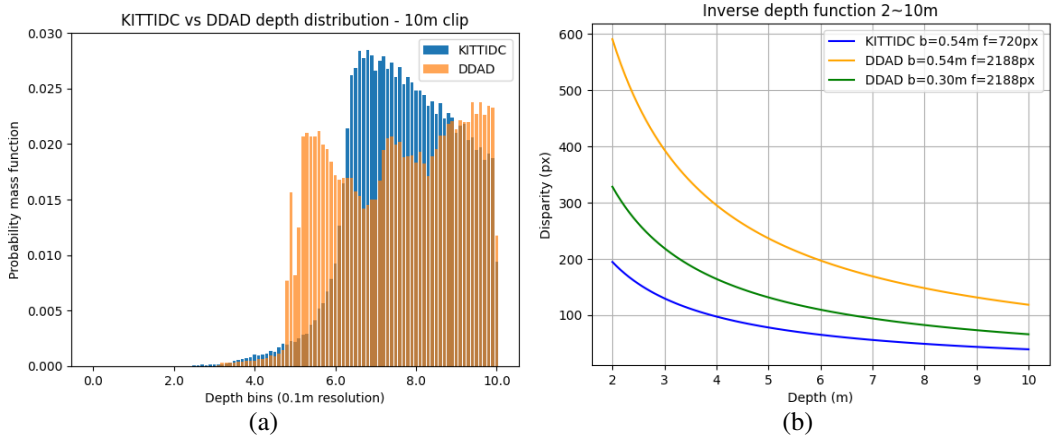

Figure VI. **DDAD [5] virtual baseline study – near depth range.** DDAD [5] and KITTIDC [14] share similar distribution of near range depth values (a). A shorter baseline aligns the disparity range of DDAD [5] to the disparity range seen during training [9] (b).

However, a reduction in baseline leads intrinsically to higher errors, that are proportional to the square of the distance between the camera and the objects. In Fig. VII (a) and Fig. VII (b) we analyze respectively the behavior of the inverse depth function in far depth bound of KITTIDC [14] (*i.e.*, 80-100m) and the one in far depth bound of DDAD [5] (*i.e.*, 200-220m). The former plot qualitatively shows that DDAD [5] resolution is higher than the one of KITTIDC [14], even with a shorter baseline: this is due to the longer focal length used to acquire the former. The latter plot shows that an error of 1px in disparity estimation equals more than 20 meters of depth error, even with a larger baseline. In conclusion, the virtual baseline is an important hyperparameter that needs to be tuned precisely to achieve optimal performance on both near and far depth ranges.

17

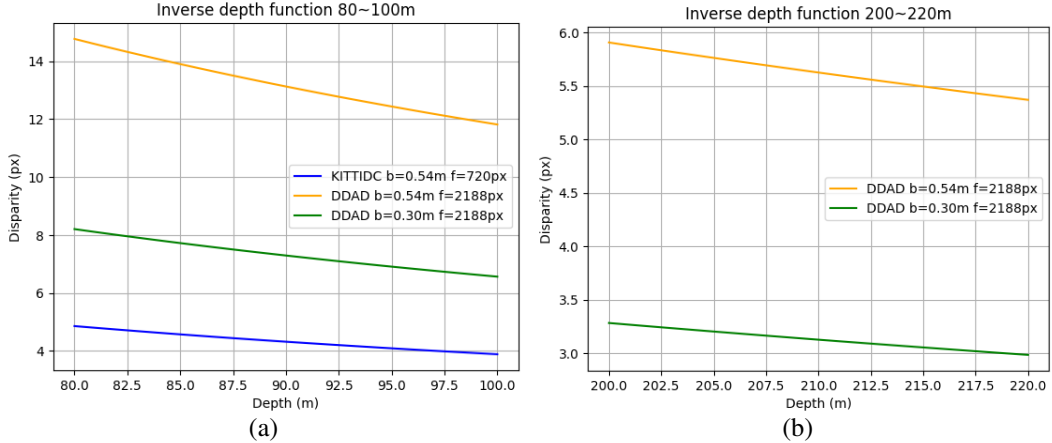

Figure VII. **DDAD [5] virtual baseline study – far depth range.** Despite the shortened baseline, DDAD [5] maintains comparable disparity resolution to KITTIDC [14], which employs a larger baseline (a). Notably, extended distances exhibit decreased accuracy, even with a larger baseline (b).

## 5. Cross domain generalization: qualitative results

We conclude by showing some qualitative examples to confirm the benefits of our method in terms of cross domain generalization. In particular, we test the performance of networks v) trained on synthetic data [9] and tested on real scenarios [5, 16]; vi) trained or finetuned indoor [11] and tested outdoor [5]; vii) trained or finetuned outdoor [14] and tested indoor [16]. We apply a $7 \times 7$ dilation to sparse depth hints and sparse ground-truth for visualization purpose.

- **v) From synthetic to real.** Fig. VIII and Fig. IX show the performance of three depth completion networks [4, 12, 19] compared to our proposal on both indoor [16] and outdoor [5] scenarios. Even if all the networks are trained on SceneFlow [9] only, in Fig. VIII they manage to reconstruct the road plane. Nonetheless, our proposal outperforms others when dealing with long distance details. Furthermore, in Fig. IX we show a challenging frame from [16] where depth hints are missing at the top of the image. In this case only our method performs adequately for the scenario, yet being not capable of producing meaningful depth values in the portion of the scene lacking any depth hints.

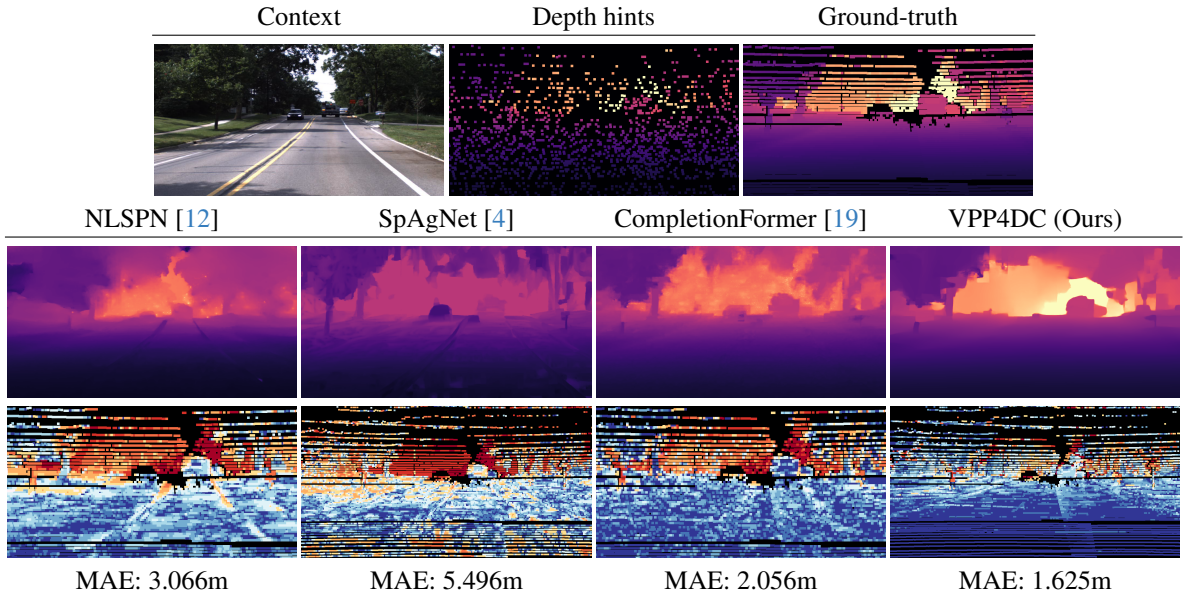

Figure VIII. **Qualitative results – DDAD [5].** All networks are trained on SceneFlow [9].

18

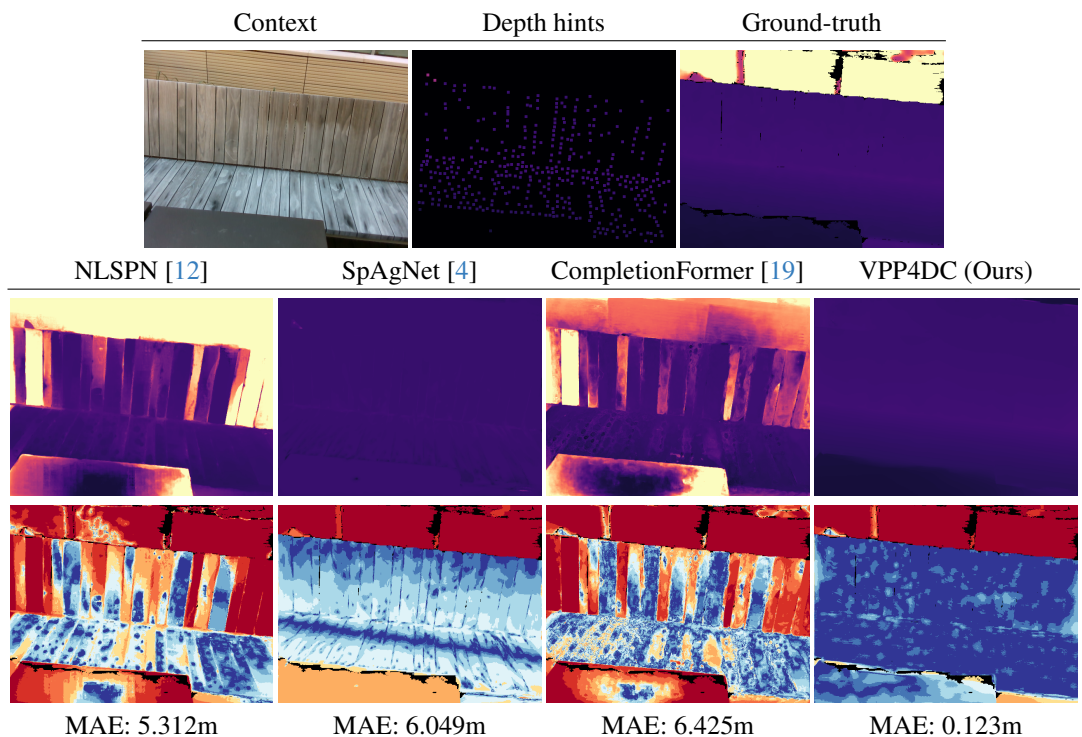

Figure IX. **Qualitative results – VOID500 [16].** All networks are trained on SceneFlow [9].

- **vi) From indoor to outdoor.** When it comes to train on indoor data and run the model outdoor, significant domain shift occurs. In Fig. X and Fig. XI we show examples on the DDAD dataset, using models trained on NYU only, or pre-trained on SceneFlow and then fine-tuned on NYU, respectively. In the former case, NLSPN and CompletionFormer produce decent results on the road surface, whereas SpAgNet completely fails at delivering any meaningful prediction. Our approach instead predicts a quite structured depth map, with minor failures. When pre-training on SceneFlow, we observe the opposite behavior: NLPSN and CompletionFormer seem unable to generalize to outdoor data, while SpAgNet can produce some meaningful depth maps, yet far from being accurate. Finally, VPP4DC can improve the results even further thanks to the pre-training process.

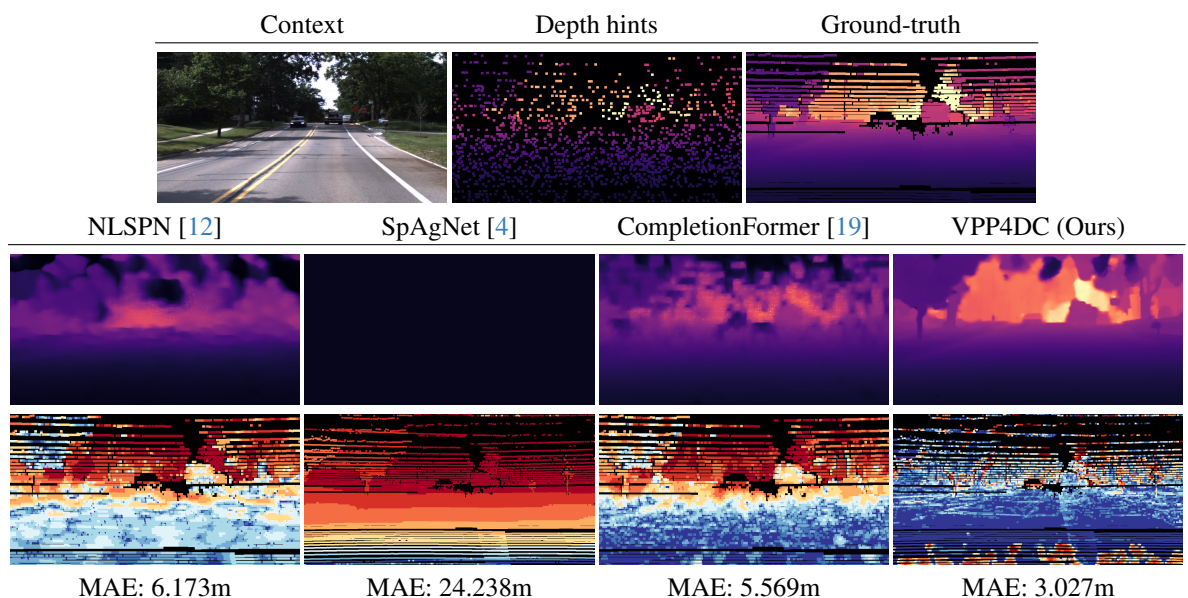

Figure X. **Qualitative results – DDAD [5].** All networks are trained on NYU [11].

19

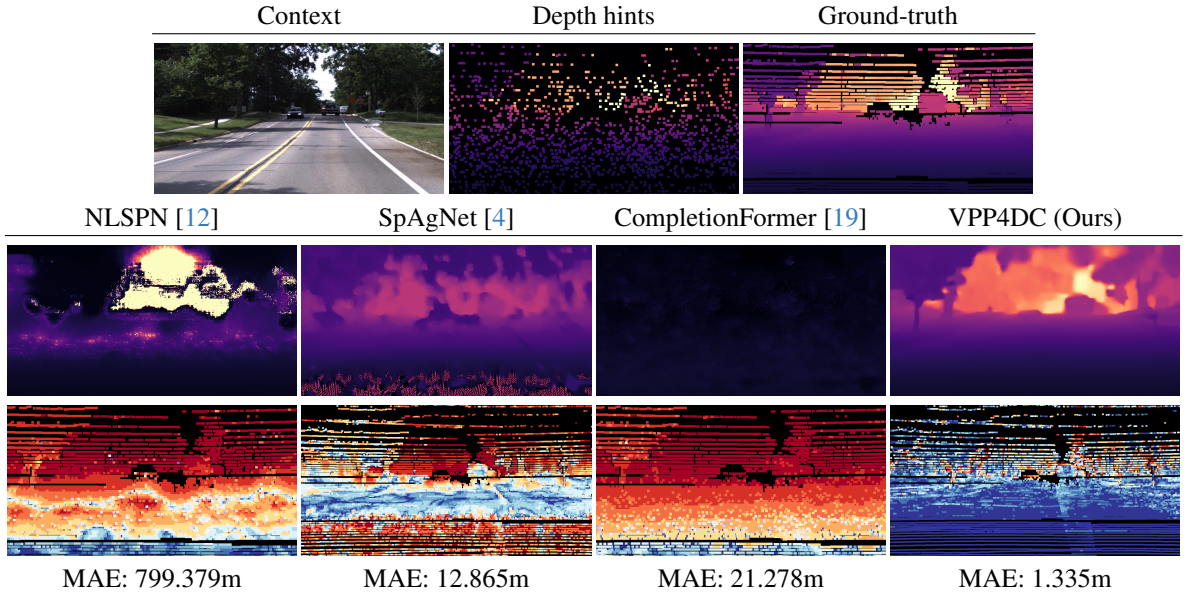

Figure XI. **Qualitative results – DDAD [5].** All networks are pretrained on SceneFlow [9] and fine-tuned on NYU [11].

- **vii) From outdoor to indoor.** Finally, we consider the case complementary to the previous one – *i.e.*, with models trained outdoor and tested indoor. In Fig. XII and Fig. XIII we show examples on the VOID dataset, using models trained on KITTI only, or pre-trained on SceneFlow and then fine-tuned on KITTI, respectively. In the former case, NLSPN and CompletionFormer completely fail, whereas SpAgNet can predict a depth map that is reasonable to some extent. Our approach instead predicts very accurate results on regions covered by depth hints, yet failing where these are absent. When pre-training on SceneFlow, NLPSN and CompletionFormer can improve their results, yet being far from the results yielded by VPP4DC without any pre-training. Eventually, if SceneFlow pre-training takes place, our method can further improve its results.

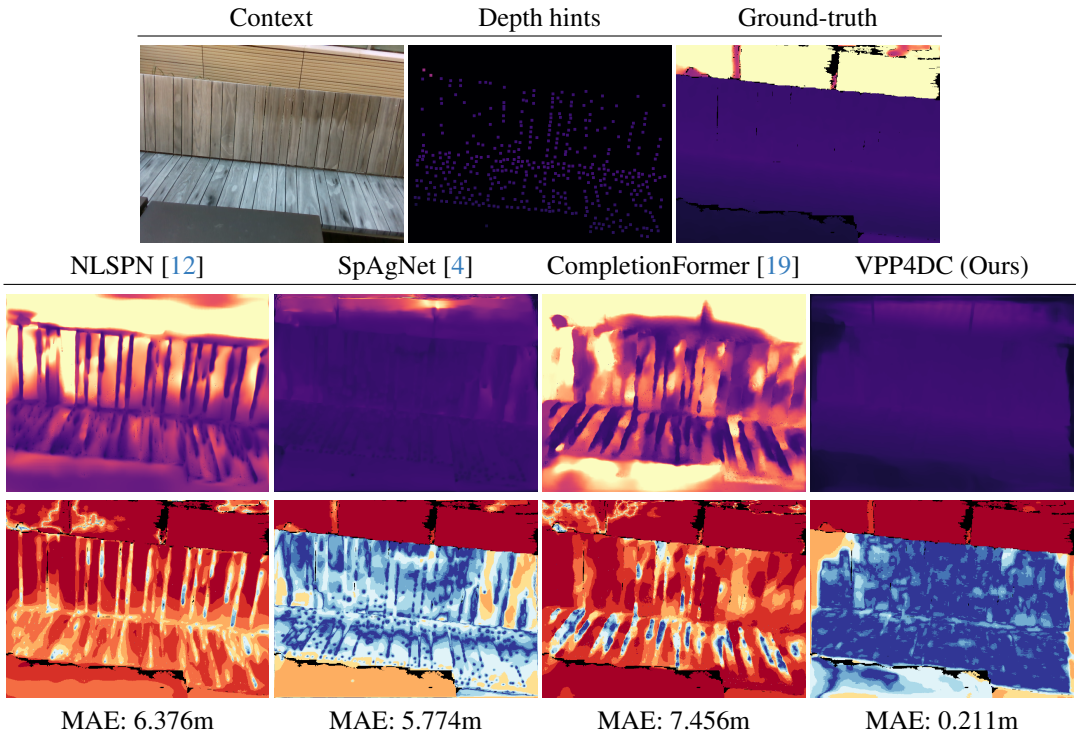

Figure XII. **Qualitative results – VOID500 [16].** All networks are trained on KITTI [14].

20

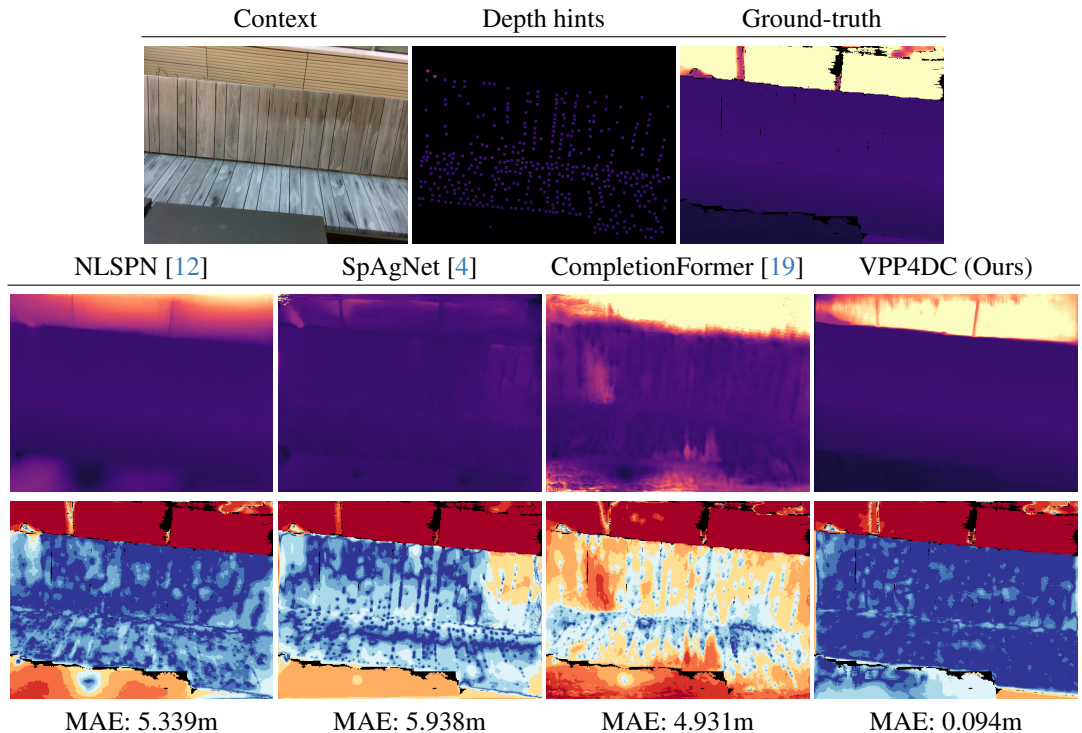

Figure XIII. **Qualitative results – VOID500 [16].** All networks are pretrained on SceneFlow [9] and finetuned on KITTI [14].

21


}

\end{document}